%% file: 1_manuscript.tex
\documentclass[lettersize,journal]{IEEEtran}

\usepackage{algorithm, algorithmic}
\usepackage{amsmath,amsfonts,amssymb, array}
\usepackage{booktabs, cite, dutchcal, epsfig}
\usepackage{forest, float, framed, graphicx}
\usepackage{import}
\usepackage{longtable,latexsym,lscape}
\usepackage{msc,multirow,multicol}
\usepackage{rotating, setspace, soul, stfloats}
\usepackage[caption=false,font=normalsize,labelfont=sf,textfont=sf]{subfig}
\usepackage{textcomp,times,tikz}
\usepackage[flushleft]{threeparttable}
\usepackage{url}
\usepackage{verbatim}
\usepackage{xcolor,xspace}
\newcommand{\etal}{\textit{et al.}}
\begin{document}

\title{MT-ProtBERT: Multi-task Learning ProtBERT for Intrinsically Disordered Proteins Classification with Scarce Data}

\author{Jian Sun,~\IEEEmembership{Member,~IEEE}, Kingshuk Ghosh, \\Lilianna Houston, Mohammad H. Mahoor~\IEEEmembership{Senior Member,~IEEE}
\thanks{Corresponding Author: Mohammad H. Mahoor.}
\thanks{Jian Sun is with the Department of Computer Science, Ritchie School of Engineering and Computer Science, University of Denver, 2155 E Wesley Ave, Denver, CO 80210,
also with the Department of Surgery, David Geffen School of Medicine at the University of California, Los Angeles, 10833 Le Conte Ave, Los Angeles, CA 90095. (Email: JianSun@mednet.ucla.edu; Home Page: https://jiansfoggy.github.io/sj-home.github.io//)}
\thanks{Mohammad H. Mahoor is with the Department of Computer Science, Ritchie School of Engineering and Computer Science, University of Denver, 2155 E Wesley Ave, Denver, CO 80210. (Email: mohammad.mahoor@du.edu; Home Page: http://mohammadmahoor.com)}
\thanks {Lilianna Houston is in the Department of Physics and Astronomy, at the University of Denver, CO 80210}
\thanks{Kingshuk Ghosh is in the Department of Physics and Astronomy, at the University of Denver, CO 80210 and also affiliated with the Molecular and Cellular Biophysics program at the University of Denver}
}

\markboth{Preprint}%
{Sun \MakeLowercase{\textit{et al.}}: MT-ProtBERT}


\maketitle

\begin{abstract}
Intrinsically disordered proteins (IDPs) differ from folded proteins in that they are dynamic, lack a stable three-dimensional conformation, and have low sequence similarity between similar proteins. 
The conformational heterogeneity of IDPs — while beneficial for their diverse functions — limits the use of traditional experimental tools to determine their conformation. 
The experimental difficulty, along with low sequence similarity, results in {\it {data scarcity}}, and makes it difficult to classify/detect IDPs that are similar or dissimilar, a task relevant to understand biology and evolution. 
We address this challenge using Multi-task ProtBERT (MT-ProtBERT), a multi-task extension of ProtBERT tailored for low-data regimes. 
MT-ProtBERT integrates Dynamic Window Masking, a Multi-Scale 1D Convolutional classifier (MS-Conv1D), and auxiliary objectives that jointly optimize masked language modeling and biochemistry-informed tasks. 
We evaluate this framework on two tasks under limited data: (i) phosphorylation site prediction (S/T/Y) in short sequences and small datasets, and (ii) protein compaction prediction on two small datasets (684 and 530 sequences), including sequences comparable in length to typical disordered regions. 
MT-ProtBERT consistently outperforms PARROT, an RNN-based IDP-specific model, across all tasks. 
These results demonstrate that combining self-supervised and biochemistry-informed tasks, and multi-scale learning enables robust modeling of unstructured proteins under data scarcity.
\end{abstract}


\section{Introduction}\label{sec:1}

\IEEEPARstart{D}{etecting} functionally similar or dissimilar proteins is fundamental in biology and biotechnology. 
Functional classification is typically done by aligning the structure or sequence of an unknown protein with that of a protein with known function~\cite{S1_44_large_2013}. 
This approach has been largely successful for proteins that have a unique folded structure. 
However, intrinsically disordered proteins (IDPs) are not amenable to this for two reasons. 
First, IDPs do not possess a stable three-dimensional structure. 
Instead, IDPs exist in an ensemble, continuously interconverting between multiple conformations. 
Second, functionally similar IDPs often share little sequence similarity\cite{S1_3_MosesPNAS_2017, S1_4_KMAD_2016}. 

Besides the classification challenge, IDPs have an additional problem of data scarcity. 
While conformational heterogeneity facilitates diverse biological functions\cite{S1_6_JMR_2010, S1_7_NatRev_2015, S1_8_JBC_2016, S1_9_Order_2021,S1_10_JProteomeRes_2007, S1_11_TIBS_2012,S1_12_Science_2009}, it also presents a challenge in experimentally determining IDP conformations. 
Experimentally determined structures for IDPs are far fewer compared to their folded counterparts. 
The lack of experimental measurements and low sequence homology, together results in \textit{data scarcity}. 
IDPs constitute 30–40\% of the human proteome \cite{S1_4_BioChem_2002, S1_5_CLSIDPs_2014}, yet the PED database contains conformations of only about ~400 IDPs from purely biochemical measurements (or in combination with simulations \cite{S1_21_PED_2021}), compared to ~250,000 folded protein structures in the PDB.

To mitigate this issue, coarse-grained simulations generate large-scale datasets \cite{S1_14_JPCL_2020, S1_15_Nature_2024} which are subsequently used to train machine learning models \cite{S1_16_JCTC_2024, S1_17_PNAS_2021, S1_18_NCS_2021, S1_45_STARLING_2026}. These physics-based models can predict IDP conformations and liquid-liquid phase separation (LLPS) propensity \cite{S1_17_PNAS_2021, S1_19_KrestenLLPS_2025}, an important function of IDPs. Nonetheless, training directly from biochemical measurements remains challenging. Additionally, the trained models do not necessarily help with functional classification other than LLPS.

A major effort in the functional classification of IDPs so far has focused on unsupervised learning. This has been supported by the emerging view that IDP sequence features correlate with conserved biophysical properties. Statistical and physics-based metrics—such as composition, patterning, and chain dimensions—have been used to predict functional similarity \cite{S1_22_ProtEngDesSelect_2019, S1_23_JMB_2020, S1_15_Nature_2024, S1_3_MosesPNAS_2017, S1_24_PNAS_2017, S1_25_elife_2019, S1_26_BJ_2021, S1_27_JPhysChemB_2025}. 

Little attention has been paid to supervised classification when IDPs have known labels, particularly in small-sample regimes (hundreds of sequences). 
In this paper we embrace the general problem of IDP classification in the small sample regime, by considering two binary classification tasks across four experiments on relatively small protein sequences and datasets to exacerbate the \textit{data scarcity} issue. In the first experiment (Experiment 1), we consider whether a given sequence will undergo phosphorylation (Task 1). Briefly, phosphorylation is a biochemical process in which the amino acids S/T/Y are modified to a negatively charged group. Phosphorylation can change the function of proteins and is biologically relevant \cite{S1_28_Nature_2015, S1_29_Science_2019}. However, not all S/T/Y amino acids are phosphorylated, thus, phosphorylation prediction is a binary task. We consider two datasets PPA and PELM~\cite{S4_1_1_PPA_2010, S4_1_2_PELM_2010} of peptides (short sequences 19 amino acids long) with a middle amino acid being S/T/Y, target site for phosphorylation. The datasets are small, for example, PPA sequences (with the central amino acid Y) include only about 1,100 samples, evenly split between phosphorylated and non-phosphorylated cases. 

In addition to predicting phosphorylation, we created an additional classification task (Task 2) of predicting protein compactness. Ensemble average end-to-end distance is a measure of global dimension of IDPs. This dimension is compared against a reference size (of an ideal protein with no interaction) to determine compact (below the reference) or expanded (above the reference) groups. This sets another binary classification task for IDPs. We test this task in Experiment 2 using a selected subset of sequences from the PELM set. Additional experiments (Experiments 3 and 4) predicting compaction (Task 2) were created by selecting slightly larger sequences (30 amino acids long) to better mimic disordered proteins from a database different from PELM\cite{S1_30_MCell_2020} and with sample size even smaller than Experiment 2.

Existing supervised IDP models primarily use CNNs \cite{S1_30_MCell_2020, S4_4_2_CapsNet_PTM_2019, S4_4_5_PhosTransfer_2020, S4_4_7_Limited_Protein_DNN_2019}, RNNs \cite{S1_36_Parrot_2021}, LSTM~\cite{S4_4_3_DeepPPSIte_2021}, or CNN-attention hybrids~\cite{S4_4_4_PhosIDN_2021, S4_4_6_EMBER_2022}. 
Phosphorylation prediction has not been explored using purely transformer-based architectures. In contrast, folded-protein studies increasingly adopt large language models~\cite{S1_13_GraphSite_2022, S1_1_ESM_2023}. 
Inspired by the analogy between protein sequences and natural language, BERT and GNN models have been applied to protein tasks~\cite{S1_37_UDSMProt_2020, S1_38_LSRMT_2023, S1_39_DeepRank-GNN_2023, S1_40_GTFE_2022, S1_46_protein_2025}, yet BERT-based approaches for IDPs remain limited \cite{S1_34_IDP-Bert_2024}. 
We therefore adopt ProtBERT~\cite{S3_2_1_ProtBERT_2021} for IDP classification.

To address \textit{data scarcity}, we propose \textbf{Multi-task ProtBERT (MT-ProtBERT)} with three components: (i) Dynamic Window Masking, which randomly masks contiguous residues of varying lengths for data augmentation; (ii) a Multi-scale 1D Convolutional classifier (MS-Conv1D) with parallel kernels to capture multi-scale features; and (iii) Multi-task Learning, integrating Masked Language Modeling (MLM) and S/T-P motif prediction as auxiliary tasks. 
MLM enhances structural representation, while S/T-P prediction embeds phosphorylation-related biochemical knowledge (Task 1).  

These modules improve robustness and outperform the RNN-based PARROT model \cite{S1_36_Parrot_2021}. 
{\it In summary, this work addresses binary classification of IDPs under data scarcity.} The main contributions are as follows:

\begin{itemize}
\item We propose \textbf{MT-ProtBERT}, a multi-task ProtBERT framework for improved classification of disordered protein sequences.
\item We prove that \textbf{multi-view MLM} outperforms single-view MLM and design a \textbf{Protein-aware Loss} to further enhance MLM performance.
\item We introduce \textbf{Dynamic Window Masking} for sequence augmentation and a \textbf{Multi-scale 1D Convolutional classifier (MS-Conv1D)} to capture multi-scale contextual features.
\item We employ \textbf{Multi-task Learning} (MLM + S/T-P prediction) to embed structural and biochemical priors, mitigating data scarcity.
\item MT-ProtBERT consistently outperforms PARROT, an RNN-based model for IDPs.
\end{itemize}

The remainder of this paper is organized as follows. Section~\ref{sec:2} reviews related work. Section~\ref{sec:3} presents the MT-ProtBERT architecture. Section~\ref{sec:4} describes datasets, experimental settings, and results. Section~\ref{sec:5} discusses implications and limitations. Section~\ref{sec:6} concludes the paper.

\section{Related Works} \label{sec:2}
This section reviews machine learning approaches for protein biophysical prediction and discusses strategies for handling data scarcity.
\subsection{Machine learning models to predict biophysical properties of proteins} \label{sec:2.1}

Protein property prediction depends on \textit{data representation}, \textit{embedding}, and \textit{feature extraction}.

\textbf{Data Representation.}  
Researchers represent proteins as sequences, graphs, or secondary structures.  
Graph-based methods encode residue interactions and geometry~\cite{S2_1_1_CLACAZy_2022, S1_13_GraphSite_2022, S1_40_GTFE_2022}, sometimes transforming secondary structures into graphs~\cite{S1_39_DeepRank-GNN_2023, S2_1_2_GraphormerDTI_2024, S1_13_GraphSite_2022}.  
Secondary structure provides geometric constraints; for example, Yuan~\etal~\cite{S1_13_GraphSite_2022} derived structures using AlphaFold2, and Chandra~\etal~\cite{S1_38_LSRMT_2023} embedded structural annotations into sequences.

However, the disordered peptides studied here lack reliable secondary structure annotations. 
We therefore adopt raw sequence representation and enhance it through masking, which increases diversity without adding structural assumptions.

\textbf{Embedding.}  
Standard NLP embeddings do not fully capture protein-specific semantics.  
Protein-tailored embeddings such as \textit{doc2vec}, \textit{Word2Vec}, and \textit{SeqVec} address this gap~\cite{prot-embed1_Yang_2018, prot-embed2_Ho_2019, prot-embed3_Hamid_2019, prot-embed4_Ofer_2021, prot-embed5_Tran_2023}.  
Embedding strategies include residue-level and $k$-mer-level representations~\cite{prot-embed2_Ho_2019, prot-embed3_Hamid_2019, prot-embed4_Ofer_2021}.  
Other methods encode biochemical properties (ESM, ESM1b, PAAC)~\cite{prot-embed1_Yang_2018, S1_43_PANDA2_2022} or structural context (Protein Graphs, Distance Maps, DSSP, RaptorX)~\cite{S1_13_GraphSite_2022, S1_42_GAT-GO_2022}.

Our datasets lack structural annotations, and ProtBERT uses residue-level tokenization. 
We therefore embed amino acids individually to maintain task compatibility.

\textbf{Feature Extraction.}  
Protein sequences contain local motifs and long-range dependencies~\cite{S1_41_ProteinBERT_2022, S1_42_GAT-GO_2022, S2_1_1_CLACAZy_2022}.  
CNNs capture local patterns, while attention-based Transformers model global context~\cite{S2_1_1_CLACAZy_2022, S1_41_ProteinBERT_2022, S1_37_UDSMProt_2020}.  
RNN-based models such as PARROT may not capture long-range dependencies~\cite{S1_36_Parrot_2021}.  

We employ ProtBERT for global modeling and integrate a multi-scale Conv1D classifier to enrich local feature extraction. 
This hybrid design balances contextual representation and structural sensitivity.

\textbf{Summary.}  
Rather than introducing new representations, we focus on extracting richer information from raw sequences while preserving interpretability and architectural simplicity.

\subsection{Data Scarcity} \label{sec:2.2}

Data scarcity arises from limited labeled samples and short sequences (19 residues in Experiments 1–2; 30 residues in Experiment 3), which constrain model capacity.

Common solutions include Transfer Learning~\cite{S2_2_1_TransLearn1_2023, S2_2_2_TransLearn2_2025}, Semi-supervised Learning~\cite{S2_2_3_SemiSL1_2024}, Few-shot Learning~\cite{S2_2_4_FewShot1_2022}, Autoencoders~\cite{S2_2_5_VAE1_2025}, Bayesian Methods~\cite{S2_2_6_Bayes1_2023}, Data Augmentation~\cite{S2_2_4_FewShot1_2022, S2_2_5_VAE1_2025}, Self-Supervised Learning~\cite{S2_2_7_SSL1_2020}, Physics-informed Modeling~\cite{S2_2_8_PhyInform1_2022}, Multi-task Learning~\cite{S2_2_8_PhyInform1_2022, S2_2_9_MultiTask2_2024}, and Multi-scale Learning~\cite{S2_2_10_MultiScale1_2022}.  

However, Transfer learning requires large dataset to prepare pretraining model. 
Semi-supervised and autoencoder approaches risk noisy pseudo-labels or posterior collapse. 
Few-shot and Bayesian methods increase computational cost.  

Data augmentation, physics-informed modeling, self-supervised learning, multi-task learning, and multi-scale learning better align with our setting. 
Data augmentation broadens the training distribution; physics-informed constraints inject domain knowledge; self-supervision extracts intrinsic sequence patterns; multi-task learning provides auxiliary guidance; and multi-scale learning captures features at different resolutions.

Guided by these principles, we design two auxiliary tasks: \textbf{Masked Token Prediction (MLM)} and \textbf{S/T-P Prediction}. 
MLM introduces self-supervision through masking, while S/T-P prediction encodes a phosphorylation-related motif at the sequence center. 
For downstream classification, we integrate a multi-scale Conv1D classifier to enhance global and local representation learning.

This combination of self-supervised, multi-task, and multi-scale learning forms our strategy to address data scarcity in IDP classification.

\section{Methodology} \label{sec:3}

\subsection{Multi-task ProtBERT} \label{sec:3.1}

\input{Figures/S3_1_Model_Structure}

Figure~\ref{fig: S3_1_MT-ProtBERT} shows MT-ProtBERT's pipeline. 
We apply Dynamic Window Masking twice to each input sequence to produce two masked views, encode both with a shared ProtBERT backbone, and obtain embeddings $f_{1}$ and $f_{2}$. 
We attach three heads: two MS-Conv1D classifiers for the main protein label and an auxiliary S/T-P or SCD label, and one MLM head that predicts masked tokens from both $f_{1}$ and $f_{2}$. 
Performing two MLM predictions per forward pass strengthens token-level representation learning. Below we describe each component.

\subsection{Dynamic Window Masking} \label{sec:3.2}

Dynamic Window Masking produces context-aware augmented sequences. 
For a sequence of length $L$, we pick a random start position $P$ and mask $k$ consecutive residues where $k\in\{1,2,3\}$ with probabilities $\{0.6,0.3,0.1\}$. The expected window size equals
\begin{align}
E[k] = 0.6\times1 + 0.3\times2 + 0.1\times3 = 1.5.
\label{eq: S3_2_1_exp_k}
\end{align}
We mask multiple non-overlapping regions to reach a total masking ratio of 20\%. 
This ratio balances information retention and diversity for short sequences (e.g., $L=19$). The number of masked regions follows
\begin{align}
\begin{split}
N_{Mask} &= \lceil20\%\times L\rceil/E[k] \\
         &\approx \lceil20\%\times L\rceil/1.5.
\end{split}
\label{eq: S3_2_2_num_mask}
\end{align}
For $L=19$, $N_{Mask}\approx2.67$, so we generate 2–3 masked regions. 
We repeat this masking twice to produce $t_{1}$ and $t_{2}$, providing multiple contextual views that increase training diversity, lower gradient variance, and speed convergence.

\subsection{ProtBERT Backbone} \label{sec:3.3}

ProtBERT~\cite{S3_2_1_ProtBERT_2021} uses amino acids as tokens and pre-trains on UniRef~\cite{S3_2_2_UniRef_2007}. 
It generalizes well to protein tasks (e.g., secondary structure, localization). 
We use ProtBERT as a shared encoder for $t_{1}$ and $t_{2}$ and obtain contextual embeddings $f_{1}$ and $f_{2}$ of size $[\text{batch size}, L, 768]$. 
These embeddings capture sequence semantics for downstream heads.

\subsection{Multi-scale 1D Convolutional Classifier (MS-Conv1D)} \label{sec:3.4}

\input{Figures/S3_4_MS-Conv1D}

Figure~\ref{fig: S3_4_ms_conv1d} depicts MS-Conv1D. We apply three parallel 1D convolutions with kernel sizes 2, 3, and 4 to ProtBERT embeddings. 
Each branch captures features at a different receptive field; we concatenate their outputs and feed them to a fully connected layer to predict either the main label or an auxiliary label (S/T-P or SCD).
This multi-scale, multi-branch design captures complementary local contexts and enhances global structural modeling, similar to MC-ViViT~\cite{S3_3_1_MCViViT_2024} and SSL-V3~\cite{S3_3_2_SSLV3_2025}.

\subsection{Multi-task Learning} \label{sec:3.5}

We train with auxiliary tasks to reinforce the main classification objective. 
Experiments 1–3 use MLM + S/T-P; Experiment 4 uses MLM + SCD.

\textbf{Masked Language Modeling (MLM).}  
The MLM head contains two fully connected layers: a projection from $f_{1}$ to vocabulary space ($f_{MLM1}$) and a decoder producing logits $f_{MLM2}\in[\text{batch size},L,V]$, where $V=30$ (20 amino acids + special tokens). 
We predict masked amino acids $AA_{1}$ from $f_{1}$ and $AA_{2}$ from $f_{2}$. 
Multi-view MLM backpropagates twice per pass and enforces token-level consistency. 
Eq.~\ref{eq: S3_5_1_LMLS} shows multi-view MLM loss $\mathcal{L}_{\text{multi}}$ does not exceed single-view loss $\mathcal{L}_{\text{single}}$ (proof in Appendix~\ref{sec: app2}).
\begin{align}
\begin{split}
\mathcal{L}_{\text{single}} &= \mathbb{E}_{M\sim \mathcal{M}}\mathbb{E}_{X\sim \text{DWM}(S,M)} \\
 &\:[-\log p_{\theta}(AA_{m}|h_{\theta}(X))] \\
\mathcal{L}_{\text{multi}} &= \mathbb{E}_{M\sim \mathcal{M}}\mathbb{E}_{X_{1},\dotsm, X_{K}\overset{iid}{\sim} \text{DWM}(S,M)} \\
&\:[-\log(\frac{1}{K}\displaystyle \sum^{K}_{i=1} p_{\theta}(AA_{m}|h_{\theta}(X_{i})))] \\
\mathcal{L}_{\text{multi}} &\leq \mathcal{L}_{\text{single}}
\end{split} \label{eq: S3_5_1_LMLS}
\end{align}

\textbf{S/T-P Prediction.}  
We predict whether the central residue forms (S or T)–P. 
The MS-Conv1D processes $f_{1}$ and outputs $f_{\text{STP}}\in[\text{batch size},2]$. 
This auxiliary task injects phosphorylation-related biochemical priors and helps the model learn motif-level structure. 
We keep S/T-P prediction in experiments where it does not directly apply (e.g., central Y) to preserve a consistent training scheme.

\textbf{SCD-based Binary Classification.}  
SCD correlates with IDP compaction~\cite{S3_5_1_JCP_2015,S3_5_2_JCP_2018,S3_5_3_JCP_2020,S3_5_4_AnnRev_2022,S3_5_5_PNASNexus_2024,S3_5_6_PNAS_2024}. 
We binarize SCD at threshold $-1.06$: set SCD=1 if SCD $> -1.06$, else 0. Experiment 4 replaces S/T-P with SCD prediction while keeping other settings identical, testing robustness with both relevant and irrelevant auxiliary tasks.

\subsection{Loss Function} \label{sec:3.6}

\subsubsection{Protein-aware Loss Function for MLM} \label{sec:3.6.1}

Each amino acid belongs to a chemical equivalence group following classification introduced by Pappu~\cite{S3_6_1_CEG_2022} (Table~\ref{tab: S3_6_chem_eqvl_ctgry}). We define a soft accuracy factor $q(aa|x_{i})$ that rewards exact matches and allows partial credit for chemically equivalent substitutions.
Let $\epsilon$ control the soft weight; we set $\epsilon=0.2$. Formally:
\input{Tables/S3_6_Chemical_Equivalence_Category}
\input{Tables/S3_6_symbols_for_Protein_Aware_Loss}
\begin{align}
q(aa|x_{i})&=\begin{cases} 1-\epsilon & aa = x_{i}\\
\frac{\epsilon}{|G(x_{i})|-1} & aa\in G(x_{i}), aa\notin x_{i}\\
0 & aa\notin G(x_{i}) \end{cases} \label{eq: S3_6_1_q_factor} \\
\mathcal{L}_{i}&=-\displaystyle\sum\limits_{aa\in AA}q(aa|x_{i})\log p_{\theta}(aa|\tilde{x}) \label{eq: S3_6_2_ith_protein_loss} \\
\mathcal{L}_{\text{mlm}}&=\frac{1}{|\mathcal{M}|}\displaystyle\sum\limits_{i\in \mathcal{M}}\mathcal{L}_{i} \label{eq: S3_6_3_overall_protein_loss}
\end{align}
For example, if $x_{i}=\text{D}$ and $G(x_{i})=\{\text{D},\text{E}\}$, then $q(\text{D}|\text{D})=0.8$ and $q(\text{E}|\text{D})=0.2$. 
We embed $q(aa|x_{i})$ into the cross-entropy loss to form the Protein-aware Loss for MLM.

\subsubsection{Overall Loss} \label{sec:3.6.2}

We combine MLM, auxiliary, and main losses. Let $\mathcal{L}_{\text{mlm1}}$, $\mathcal{L}_{\text{mlm2}}$ denote MLM losses for $t_{1}$ and $t_{2}$, $\mathcal{L}_{\text{stp}}$ denote the auxiliary loss, and $\mathcal{L}_{\text{main}}$ denote the main classification loss. We define:
\begin{equation}
\begin{split}
\mathcal{L}_{\text{MLM}} &= 0.5 \times (\mathcal{L}_{\text{mlm1}} + \mathcal{L}_{\text{mlm2}}) \\
\mathcal{L} &= \mathcal{L}_{MLM} + 0.1 \times (\mathcal{L}_{\text{stp}} + \mathcal{L}_{\text{main}}).
\end{split}
\label{eq: S3_6_4_loss_stage1}
\end{equation}
We scale auxiliary and main losses by 0.1 to align magnitudes. 
We emphasize MLM during early training to learn structural and semantic regularities; after epoch 25, MLM primarily stabilizes token-level representations.

Combining Dynamic Window Masking, ProtBERT, MS-Conv1D, and multi-task objectives lets MT-ProtBERT learn physicochemical and structural patterns from limited data and improve IDP classification.

\section{Experiment} \label{sec:4}
This section presents the datasets, implementation details, evaluation metrics, experimental results, and analyses used to assess the effectiveness of the proposed MT-ProtBERT model.

\subsection{Dataset} \label{sec:4.1}
We evaluate MT-ProtBERT on two benchmark phosphorylation datasets: \textbf{PPA} and \textbf{PELM}.  
\textbf{PPA} originates from the PhosphAt database and contains phosphorylation sites exclusively from Arabidopsis thaliana~\cite{S4_1_1_PPA_2010}.  
\textbf{PELM}, created by Dinkel~\etal, is derived from the Phospho.ELM database and includes experimentally verified phosphorylation sites in various animal species such as Homo sapiens and Mus musculus~\cite{S4_1_2_PELM_2010}. 

\input{Tables/S4_0_Sample_Display}

Table~\ref{tab: S4_0_data_smp} lists representative sequence samples from both datasets.
Each sequence is 19 amino acids long, centered on a target residue (S, T, or Y). 
The label indicates whether the site is phosphorylation-active (\texttt{Active}) or non-active (\texttt{Non-active}).

Both PPA and PELM contain balanced positive and negative classes (28,001 samples each), as summarized in Table~\ref{tab: S4_1_dataset_stats}. 
However, subsets with central residue T or Y contain far fewer samples than those with S, resulting in a pronounced data scarcity problem. 
For example, PPA with central Y has only 1,156 samples. 
This is a motivation to choose the dataset to test models in the low sample regime. 
This scarcity, combined with short sequence lengths, poses a challenge for deep learning models, which motivates our multi-task and multi-scale learning strategy.

\input{Tables/S4_1_Data_Distribution}

To evaluate MT-ProtBERT on a different task and small datasets, we design Experiments 2–4 for protein compaction prediction.  
We quantify compaction using the ratio $x = R_{ee}^2 / R_{ee, FRC}^2$, where $R_{ee} = \langle r_{ee}^2 \rangle^{1/2}$ is the ensemble-averaged end-to-end distance and $R_{ee, FRC}$ corresponds to a Flory Random Coil (FRC), an ideal non-interacting reference chain.  
$x>1$ indicates expansion and $x<1$ indicates collapse, yielding a binary classification.  
A previously developed physics-based machine learning model -- verified against experimental data -- was used to predict $x$ directly from sequence~\cite{S1_16_JCTC_2024}.

\textbf{PELM-684 (Experiment 2).}  
We construct a balanced subset from PELM.  
Sequences with $0.90<x<0.98$ form the compact class, excluding values near 1 to avoid ambiguous cases.  
We then sample sequences with $1.02<x<1.10$ to match this distribution as the expanded class.  
The final PELM-684 dataset contains 684 sequences evenly split between compact and expanded classes.

\textbf{AD-530 (Experiments 3 and 4).}  
We derive a second balanced dataset from 30-residue sequences generated by Erijman et al.~\cite{S1_30_MCell_2020}, originally designed to study activation domains in yeast. 
These sequence lengths better reflect intrinsically disordered regions.  
Sequences with $x<0.99$ define the compact class, and sequences with $x>1.01$ define the expanded class.  
After matching class distributions, the AD-530 dataset contains 530 sequences with equal class sizes.  
Experiment 3 evaluates compaction prediction on this unseen dataset.  
Experiment 4 uses the same AD-530 set but replaces the S/T-P auxiliary task with SCD-based binary classification (Section~\ref{sec:3.4}).

\subsection{Implementation Details} \label{sec:4.2}

All models were implemented in PyTorch. 
The input amino acid sequences were first tokenized and embedded using the ProtBERT tokenizer. 
We used an 9:1 split for training and testing, with no overlap between subsets.  
Training was conducted on an NVIDIA RTX 3090 Ti GPU using the AdamW optimizer with an initial learning rate of $1\times10^{-6}$ and batch size of 32. 
The model was trained for 30 epochs with a cyclic learning rate scheduler (triangular2 mode). 
The backbone parameters of ProtBERT were fine-tuned jointly with the multi-task heads.  

Hyper-parameters for Dynamic Window Masking were empirically set to mask 20\% of residues per sequence, with a window length expectation of $E[k]=1.5$. 
All other configurations followed the same setup across both datasets.

\subsection{Metrics} \label{sec:4.3}

We evaluate model performance using six commonly adopted metrics for binary classification in bioinformatics:  
\textbf{Accuracy}, \textbf{F1 Score}, \textbf{AUPRC} (Area Under the Precision–Recall Curve), \textbf{Sensitivity}, \textbf{Specificity}, \textbf{MCC} (Matthews Correlation Coefficient), \textbf{PLL} (Pseudo-Log-Likelihood), \textbf{MRR} (Mean Reciprocal Rank).  
Among them, MCC provides a balanced evaluation even under class imbalance, and AUPRC is particularly informative for tasks with limited positive samples.
PLL measures the average log-likelihood of the ground-truth amino acids over all masked positions, ranging $(-\infty, 0]$. 
MRR emphasizes how early the correct amino acid appears in the ranked list, ranging $(0,1]$. Higher PLL and MRR indicate better MLM performance.

\subsection{Protein Phosphorylation Prediction} \label{sec:4.4}

\input{Tables/S4_2_Tsk1_Exp1_Results}

Table~\ref{tab: Tsk1_Exp1_results} compares MT-ProtBERT with PARROT and MusiteDeep on PPA and PELM.  
For subsets with central residue S, MT-ProtBERT achieves accuracy comparable to or slightly higher than existing methods.  
On smaller subsets, performance gains become substantial. 
For PPA with central residue T (2,722 sequences) and Y (1,156 sequences), MT-ProtBERT improves mean accuracy by at least 6.6\% and 9.6\% over PARROT and MusiteDeep, respectively. 
It also achieves the highest MCC across all subsets.  

MT-ProtBERT generally outperforms competing methods on most metrics with the exception of AUPRC (higher in PARROT on PPA) and Sensitivity (higher in MusiteDeep on PELM). 
However, MusiteDeep’s high Sensitivity accompanies low Specificity, indicating biased predictions. 
Similarly, PARROT’s higher AUPRC often coincides with reduced Sensitivity and/or Specificity.  
These results show that integrating Dynamic Window Masking, MS-Conv1D, and multi-task learning improves performance while maintaining balanced metrics under data scarcity.

Several other phosphorylation predictors exist~\cite{S4_4_1_Transphos_2022, S4_4_2_CapsNet_PTM_2019, S4_4_3_DeepPPSIte_2021, S4_4_4_PhosIDN_2021, S4_4_5_PhosTransfer_2020, S4_4_6_EMBER_2022, S4_5_1_review_2025}.  
However, most use different datasets, sequence lengths, or incomplete evaluation metrics, preventing direct comparison.  
CapsNet PTM~\cite{S4_4_2_CapsNet_PTM_2019}, PhosIDN~\cite{S4_4_4_PhosIDN_2021}, PhosTransfer~\cite{S4_4_5_PhosTransfer_2020}, and EMBER~\cite{S4_4_6_EMBER_2022} employ datasets distinct from PPA/PELM and different input lengths, and some do not report accuracy or MCC.  
TransPhos~\cite{S4_4_1_Transphos_2022}, DeepPPSIte~\cite{S4_4_3_DeepPPSIte_2021}, and Limited Protein DNN~\cite{S4_4_7_Limited_Protein_DNN_2019} use PPA and PELM, but TransPhos and DeepPPSIte rely on large datasets, while Limited Protein DNN targets relatively small samples with sequences of only 9 residues—much shorter than typical disordered regions and the 19-residue sequences used in this study.

\subsection{Protein Compaction Prediction} \label{sec:4.5}

To evaluate generalization beyond phosphorylation, we introduced protein compaction prediction as an additional task.
Experiment 2 uses the small PELM-684 dataset (Section~\ref{sec:4.1}).  
Experiments 3 and 4 use the AD-530 dataset, which contains 30-residue sequences outside PPA and PELM.
Experiment 3 tests performance on sequences of different length comparable to intrinsically disordered regions.  
We compare MT-ProtBERT with PARROT and its optimized variant (Table~\ref{tab: Tsk2_Exp2_4_results}).  
In Experiments 2 and 3, MT-ProtBERT consistently surpasses PARROT in distinguishing compact and expanded sequences, improving accuracy by up to 13\% and substantially increasing MCC.  

In Experiment 4, we replace S/T-P with a different auxiliary task. We convert sequence charge decoration (SCD) into a binary classification task using a threshold (see section 3.5). 
With SCD supervision, MT-ProtBERT achieves performance comparable to Experiment 3 and remains superior to PARROT.  

These results show that MT-ProtBERT generalizes beyond phosphorylation and does not depend on S/T-P as the only auxiliary task. 
This also shows the framework can be readily used to other biophysically grounded objectives.

\input{Tables/S4_3_Tsk2_Exp2_4_Results}

\subsection{MLM, S/T-P, and SCD Prediction} \label{sec:4.6}

\input{Tables/S4_4_MLM_STP_SCD_Results}

\textbf{Masked Language Modeling (MLM).}  
Table~\ref{tab: mlm_stp_scd_results} reports MLM performance.  
MT-ProtBERT achieves higher Top-1 and Top-5 accuracy on PELM than on PPA, reflecting the effect of sample size.  
Top-1 accuracy ranges from 46.58\%–52.51\% on PELM and 40.49\%–49.69\% on PPA.  
All PLL values remain below -2, and MRR stays between 45\% and 58.5\%, indicating stable training.  

Within PPA, accuracy and MRR follow S $>$ T $>$ Y, consistent with subset size; PLL shows the same order in both PPA and PELM. 
Larger datasets therefore improve masked residue recovery and strengthen structural priors.  
Top-5 accuracy substantially exceeds Top-1, showing that the model often ranks the correct residue among its top candidates even when it misses the top prediction.

In compaction prediction (Experiment 2), MLM achieves 50.45\% soft Top-1 accuracy.
When the dataset shrinks from 684 to 530 samples (Experiments 3 and 4), Top-1 accuracy drops to 28.86\% and 28.66\%.  
Thus, task type, dataset distribution, and sample size directly influence auxiliary-task performance.

\textbf{S/T-P Prediction.}  
S/T-P prediction encodes a biologically relevant phosphorylation motif (e.g., CDK-recognized S/T-P).  
MT-ProtBERT achieves near-perfect accuracy in identifying whether the central residue is S/T and the adjacent residue is P, confirming that the model captures biochemical motifs.  
For compaction prediction (Experiments 2–3), S/T-P accuracy reaches 100\%, indicating that this auxiliary task is relatively simple.

\textbf{SCD-based Binary Prediction.}  
Experiment 4 replaces S/T-P with SCD-based binary classification.  
The auxiliary accuracy reaches 72.07\%, lower than S/T-P in Experiments 2–3.  
However, Experiment 4 improves compaction prediction accuracy by 3\% over Experiment 3.  
Because SCD directly relates to charge patterning and compaction, this auxiliary task aligns more closely with the main objective and yields stronger synergy.

\subsection{Ablation Study} \label{sec:4.7}

\input{Figures/S4_7_Tsk1_Exp1_Abl_plot}

We evaluated each module by selectively removing or modifying components. 
Figures~\ref{fig: S4_7_tsk1_exp1_abl_plot}–\ref{fig: S4_7_tsk2_exp4_abl_plot} and Tables~\ref{tab: SApp_1_Abl_Tsk1_Exp1}–\ref{tab: SApp_1_Abl_Tsk2_Exp4} (Appendix~\ref{sec: app}) summarize the results.

\textbf{Multi-position Dynamic Window Masking.}  

Multi-position masking places non-overlapping masks at several locations instead of a single site, exposing the model to diverse local contexts.  
In Task 1 (Experiment 1), it improves accuracy and MCC by up to 3.5\% and 7.6\%, respectively, on small subsets such as PPA (AA=Y), with similar trends across other subsets.  
In Task 2 (Experiments 2–4), multi-position masking consistently outperforms single-position masking on small datasets (see Figures \ref{fig: S4_7_tsk2_exp2_abl_plot}, \ref{fig: S4_7_tsk2_exp3_abl_plot} and \ref{fig: S4_7_tsk2_exp4_abl_plot}).
Multiple masked views, therefore, enhance robustness under limited data.

\input{Figures/S4_7_Tsk2_Exp2_Abl_plot}

\textbf{MS-Conv1D.}  

Replacing MS-Conv1D with single-scale convolution degrades performance across subsets.  
In Task 1, MCC drops by at least 1.1\% for PPA (AA=T) and PELM (AA=Y). 
Gains in MCC generally exceed those in accuracy, and Task 2 shows the same pattern (see Figures \ref{fig: S4_7_tsk2_exp2_abl_plot}, \ref{fig: S4_7_tsk2_exp3_abl_plot}, and \ref{fig: S4_7_tsk2_exp4_abl_plot}).  
Multi-scale convolution captures complementary spatial dependencies and supports hierarchical protein feature learning.

\textbf{MLM Objective.}  
Adding MLM consistently improves both accuracy and MCC, with larger gains in small datasets.  
For example, in Task 1 (PELM, AA=Y), MLM increases accuracy by 0.29\%. 
MCC gains reach 5.3\% (PPA, AA=Y) and 2.9\% (PELM, AA=T).  
In Experiment 2, MLM improves accuracy by 2.7\% and MCC by 7.0\%.  
These results show that MLM strengthens contextual encoding and token-level semantic consistency.

\input{Figures/S4_7_Tsk2_Exp3_Abl_plot}

\textbf{S/T-P or SCD Prediction.}  

Adding S/T-P prediction yields modest but consistent gains, especially in MCC for small subsets.  
In Task 1 (PPA and PELM, AA=Y), MCC increases by 7.5\% and 0.2\%, while accuracy improves by 0.3–4.4\%.  
In Experiments 2–3, accuracy improves by 1.2–2.6\% and MCC by 1.9–5.4\%.  
This auxiliary task guides the model toward biochemically meaningful representations.  

In Experiment 4, SCD prediction produces larger gains—6.4\% in accuracy and 13.8\% in MCC—indicating stronger relevance to protein compaction than S/T-P prediction (Figure \ref{fig: S4_7_tsk2_exp4_abl_plot}).

\textbf{Multi-task Learning.}  

Combining MLM and S/T-P under the multi-task framework yields clear gains.  
In Experiment 1, full MT-ProtBERT improves accuracy by 0.4–4\% and MCC by 0.5–7\% compared with removing both MLM and S/T-P, except for PELM (AA=S), where performance remains similar.  
In protein compaction prediction (Experiments 2–4), multi-task learning improves accuracy by 4–7\% and MCC by 8.5–15.8\%.  
These results confirm that structural (MLM) and biochemical (S/T-P) priors act synergistically.

\textbf{Multi-view MLM.}  

Multi-view MLM consistently improves accuracy and MCC across experiments, except for PELM (AA=S), where performance remains near-optimal.  
Overall, multi-view MLM plays a critical role in MT-ProtBERT.

\textbf{Protein-aware Loss Function.}  

Protein-aware Loss enables best or near-best performance in Task 1, including nearly 2\% accuracy improvement on PPA (AA=Y).  
In Task 2, gains exceed 4\% in Experiments 3 and 4.  
These results justify integrating Protein-aware Loss into MT-ProtBERT.

\input{Figures/S4_7_Tsk2_Exp4_Abl_plot}

\subsection{Data Analysis and Summary} \label{sec:4.8}

The ablation study shows that Dynamic Window Masking, MS-Conv1D, and Multi-task Learning consistently improve accuracy and MCC, especially on small datasets. 
ProtBERT performs well on large subsets (e.g., PELM, central residue S, 31K samples) but degrades sharply in low-data settings such as PPA (central residue Y).  

In contrast, MT-ProtBERT maintains stable and strong performance across all subsets, demonstrating robustness to limited data and short sequences. 
It achieves a favorable balance between accuracy and model complexity, making it well-suited for phosphorylation and compaction prediction under realistic small-sample constraints.

\section{Discussion} \label{sec:5}

This study proposes MT-ProtBERT, a multi-task framework for predicting phosphorylation (Task 1) and compaction (Task 2) of IDPs when data is scarce. By integrating self-supervised and protein-informed objectives, MT-ProtBERT consistently outperforms state-of-the-art baselines such as PARROT.  
On the most data-scarce subset (PPA, central residue Y, 1,156 samples), it achieves 65.57\% accuracy and 31.42\% MCC in Task 1, exceeding other models by nearly or over 10\%. 
Strong performance is also observed in Task 2. 
These results indicate that the proposed modules enable ProtBERT to learn more generalizable and physically meaningful sequence representations for diverse IDP properties.

\subsection{Multi-task Learning Benefits} \label{sec:5.1}

The performance gains stem from synergistic auxiliary tasks. 
Masked Language Modeling (MLM) captures sequence structure and contextual dependencies, while multi-view MLM strengthens supervision and feature extraction. 
The S/T-P prediction task embeds biochemical priors by emphasizing phosphorylation motifs (e.g., S/T followed by P). 
Together, these complementary signals regularize the backbone, reduce overfitting, and enhance robustness in low-data regimes.  

However, improvements over ProtBERT diminish on large datasets (e.g., PELM, AA=S), where sufficient supervision allows full task specialization. 
Auxiliary constraints mainly benefit small-sample settings; thus, MT-ProtBERT is best viewed as a robustness-enhancing architecture for data-scarce scenarios rather than a universal replacement in high-resource cases.

\subsection{Role of Dynamic Window Masking and MS-Conv1D} \label{sec:5.2}

Dynamic Window Masking augments training by masking multiple random regions, increasing contextual diversity without external data synthesis. 
The Multi-scale 1D Convolutional (MS-Conv1D) classifier extracts hierarchical features across different receptive fields, capturing both local motifs and long-range dependencies consistent with protein structural hierarchy.

\subsection{Impact of Data Volume and Distribution} \label{sec:5.3}

Model performance strongly correlates with sample size. 
ProtBERT performs best on large subsets (e.g., PELM, central residue S, 31K samples, 77.45\% accuracy) but degrades markedly in small datasets (e.g., PPA, Y). 
MT-ProtBERT mitigates this drop by leveraging auxiliary objectives to extract additional structural information, compensating for limited labeled data.

\subsection{Impact of Protein-aware Loss function} \label{sec:5.4}

The Protein-aware Loss incorporates chemical equivalence categories (Table~\ref{tab: S3_6_chem_eqvl_ctgry}), embedding biochemical knowledge into optimization. 
Compared with the “No Soft Acc” setting, it yields up to 10\% accuracy improvements (Figures~\ref{fig: S4_7_tsk1_exp1_abl_plot}, \ref{fig: S4_7_tsk2_exp2_abl_plot}, \ref{fig: S4_7_tsk2_exp3_abl_plot}, and \ref{fig: S4_7_tsk2_exp4_abl_plot}), leading to more biologically consistent predictions.

\subsection{Model Efficiency and Feasibility} \label{sec:5.5}

MT-ProtBERT maintains architectural simplicity and computational efficiency while outperforming prior methods. 
It generalizes across datasets without external handcrafted features or large-scale fine-tuning (Table~\ref{tab: Tsk1_Exp1_results}), demonstrating the adaptability of BERT-based models in IDP research and providing a template for other molecular tasks with limited data.

\subsection{Addressing Data Scarcity through Self-supervision} \label{sec:5.6}

Data scarcity is alleviated through intrinsic self-supervision (MLM) and biochemical auxiliary tasks (S/T-P prediction). 
Rather than relying solely on annotations, the model exploits sequence structure and residue relationships as implicit supervision. 
Experiment 4 further supports this strategy by incorporating a physics-based metric (SCD) as an auxiliary task for compaction prediction, improving generalization in small datasets.

\subsection{Limitations and Future Directions} \label{sec:5.7}

Three limitations remain.  
First, MLM Top-1 accuracy is relatively low due to short sequence length (19 residues), limiting effective masking and contextual reconstruction.  
Second, S/T-P prediction rapidly saturates (100\% accuracy), indicating coarse supervision. 
Future work could incorporate richer physical descriptors such as hydrophobicity, solvent-accessible surface area, torsional angles, or conformation-related metrics (e.g., SCD~\cite{S3_5_1_JCP_2015}, SCDM, SHDM~\cite{S3_5_4_AnnRev_2022}). 
Contrastive learning or Conditional Variational Autoencoders (CVAE) may further enhance representation diversity.  
Third, current experiments focus on equal-length sequences ($\leq$ 30 residues), whereas functionally similar IDPs may vary in length. 
Extending MT-ProtBERT to handle variable and longer sequences is therefore essential.

\subsection{Summary} \label{sec:5.8}

MT-ProtBERT offers an effective and interpretable framework for IDP classification when data is scarce. 
By integrating self-supervised, multi-scale, and biology-informed learning, it bridges sequence modeling and biochemical priors, providing methodological insights for data-efficient deep learning in molecular biophysics.

\section{Conclusion} \label{sec:6}

In this work, we developed \textbf{MT-ProtBERT}, a multi-task learning framework for disordered protein classification in the limit of small sample size. The model integrates several complementary modules—including Dynamic Window Masking, ProtBERT backbone, MS-Conv1D, and two auxiliary heads for MLM and another sequence-based (S/T-P or SCD) classification task.  
Through this design, MT-ProtBERT effectively captures both global biochemical patterns and local structural motifs, achieving strong generalization even in short and sparse protein sequences such as PPA and PELM. 

\subsection{Future Work} \label{sec:6.1}
Building upon MT-ProtBERT, future research will explore several new directions.  
First, we plan to design a \textbf{Self-Adaptive ProtBERT} capable of dynamically adjusting its learning behavior based on data sufficiency—performing robustly across both data-abundant and data-scarce regimes. 
For abundant samples, the Self-Adaptive model should only use the main task, while for the data scarce case, it would add auxiliary tasks.  
Second, to better learn sequence structure, we aim to replace MLM task by \textbf{Conditional Variational Autoencoder (CVAE)}, reconstructing token embeddings from ProtBERT outputs.  
Third, we will investigate contrastive learning strategies to alleviate posterior collapse in CVAE-based architectures and to formally establish the mathematical link between latent regularization and representation stability~\cite{S1_22_ProtEngDesSelect_2019, S1_23_JMB_2020, S1_15_Nature_2024, S1_3_MosesPNAS_2017, S1_24_PNAS_2017, S1_25_elife_2019, S1_26_BJ_2021, S3_5_4_AnnRev_2022, S1_27_JPhysChemB_2025}.

\subsection{Broader Impact} \label{sec:6.2}
Overall, MT-ProtBERT demonstrates that coupling multi-task learning, and self-supervised objectives offers a scalable path forward for datasets with small sample size, typical in modeling IDPs.
Beyond phosphorylation site prediction, the principles established here can generalize to a wide range of molecular sequence modeling tasks—such as protein–protein interaction, detailed modeling of IDP conformation ensemble, and mutation impact analysis—where labeled data are often limited.  
By combining interpretability, efficiency, and adaptability, MT-ProtBERT contributes to the broader goal of developing robust, data-efficient deep learning systems for computational biology and protein science.

\section*{Acknowledgments}
We acknowledge support from NIH R01GM138901.

\bibliographystyle{IEEEtran}
\bibliography{cas-refs}

\appendices
\section{Ablation Study Results} \label{sec: app}

\input{Tables/SApp_Ablt_Tsk1_Exp1}

\input{Tables/SApp_Ablt_Tsk2_Exp2}

\input{Tables/SApp_Ablt_Tsk2_Exp3}

\input{Tables/SApp_Ablt_Tsk2_Exp4}

Tables~\ref{tab: SApp_1_Abl_Tsk1_Exp1}, \ref{tab: SApp_1_Abl_Tsk2_Exp2}, \ref{tab: SApp_1_Abl_Tsk2_Exp3}, and~\ref{tab: SApp_1_Abl_Tsk2_Exp4} present the detailed results from the ablation studies conducted on both PPA/PELM and new datasets.  
These results comprehensively verify the contribution of each proposed component within the MT-ProtBERT architecture.  
Specifically, the integration of \textbf{Dynamic Window Masking}, \textbf{MS-Conv1D}, \textbf{Multi-view MLM}, \textbf{ProtBERT-aware Loss function}, and \textbf{Multi-task Learning} consistently leads to superior performance across most metrics, including Accuracy, F1 Score, and MCC.

The experiments reveal several key observations:

\begin{itemize}
    \item \textbf{Dynamic Window Masking} provides multiple contextual views of short protein sequences, enhancing the model’s capacity to learn both local residue-level and global structural information.
    \item \textbf{MS-Conv1D} introduces multi-scale convolutional filters that enrich token embeddings and improve feature generalization, particularly beneficial for data-scarce subsets such as PPA with AA=Y and new dataset.
    \item \textbf{Multi-view MLM} (use both $AA_{1}$ and $AA_{2}$) provides extra supervision.
    \item \textbf{Multi-task Learning} (combining MLM and S/TP prediction) injects diverse supervision signals that help stabilize the ProtBERT encoder, reduce overfitting, and guide the model toward learning physics-aware and semantically consistent representations.
    \item \textbf{ProtBERT-aware Loss function} embeds biology knowledge into the model.
\end{itemize}

Overall, the ablation results clearly demonstrate that the full MT-ProtBERT configuration achieves the most balanced and robust performance across different amino acid types and datasets.  
This validates the necessity of each component in addressing both \textbf{protein compaction and size prediction} and the underlying \textbf{data scarcity challenge}.

\section{The Effectiveness of Multi-view MLM} \label{sec: app2}

\input{2_Appendix_Prove_2AA_is_better}

\section*{Biography Section}

\begin{IEEEbiography}[{\includegraphics[width=1in,height=1.25in, clip,keepaspectratio]{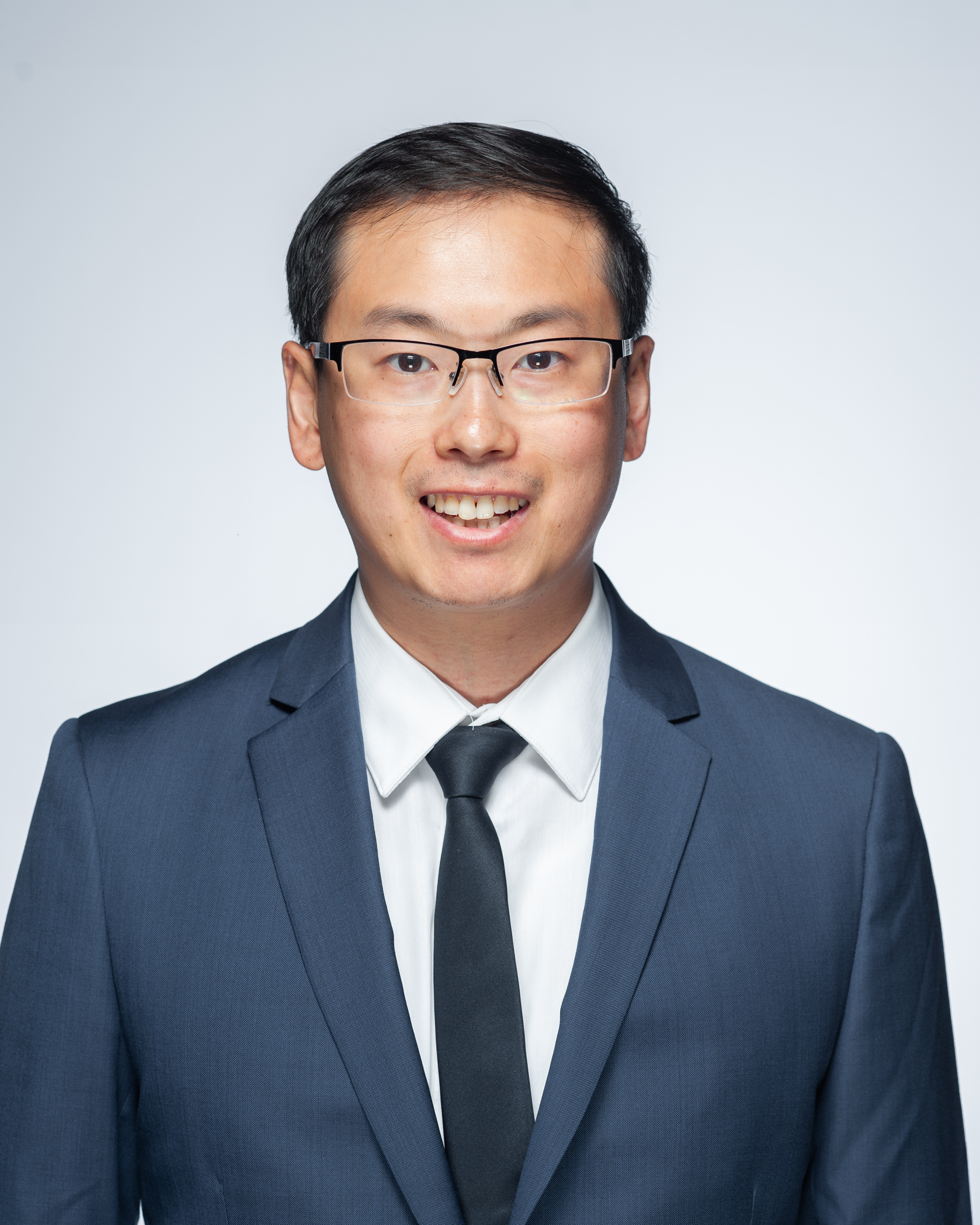}}]{Jian Sun} is a postdoctoral scholar at the University of California, Los Angeles, focusing on medical AI. 
Jian received the Ph.D. in Computer Science at the University of Denver with an emphasis on data-related issues in machine learning in 2026, received the M.S. in Statistics from the George Washington University in 2017, received B.S. in Mathematics and Applied Mathematics from Shandong Agricultural University in 2014.
Jian's research interests include Computer Vision, NLP, and Speech Analysis. 
\end{IEEEbiography}


\begin{IEEEbiography}[{\includegraphics[width=1in,height=1.25in, clip,keepaspectratio]{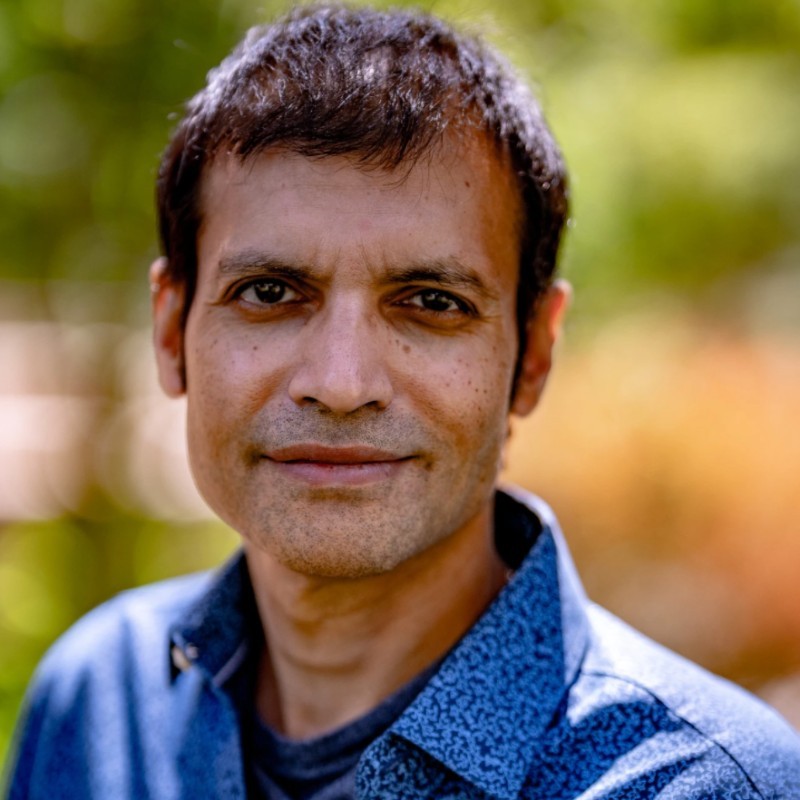}}]{Kingshuk Ghosh} is a Professor in the Department of Physics and Astronomy at the University of Denver working on theoretical biophysics. 
Ghosh received his Ph.D. in Physics from the University of Massachusetts, Amherst in 2003 and worked as a Post-doctoral scholar at the University of California, San Francisco. 
Ghosh groups works on building physics based models of disordered proteins, gene networks and other complex problems in Biophysics.
\end{IEEEbiography}

\begin{IEEEbiography}[{\includegraphics[width=1in,height=1.25in,clip,keepaspectratio]{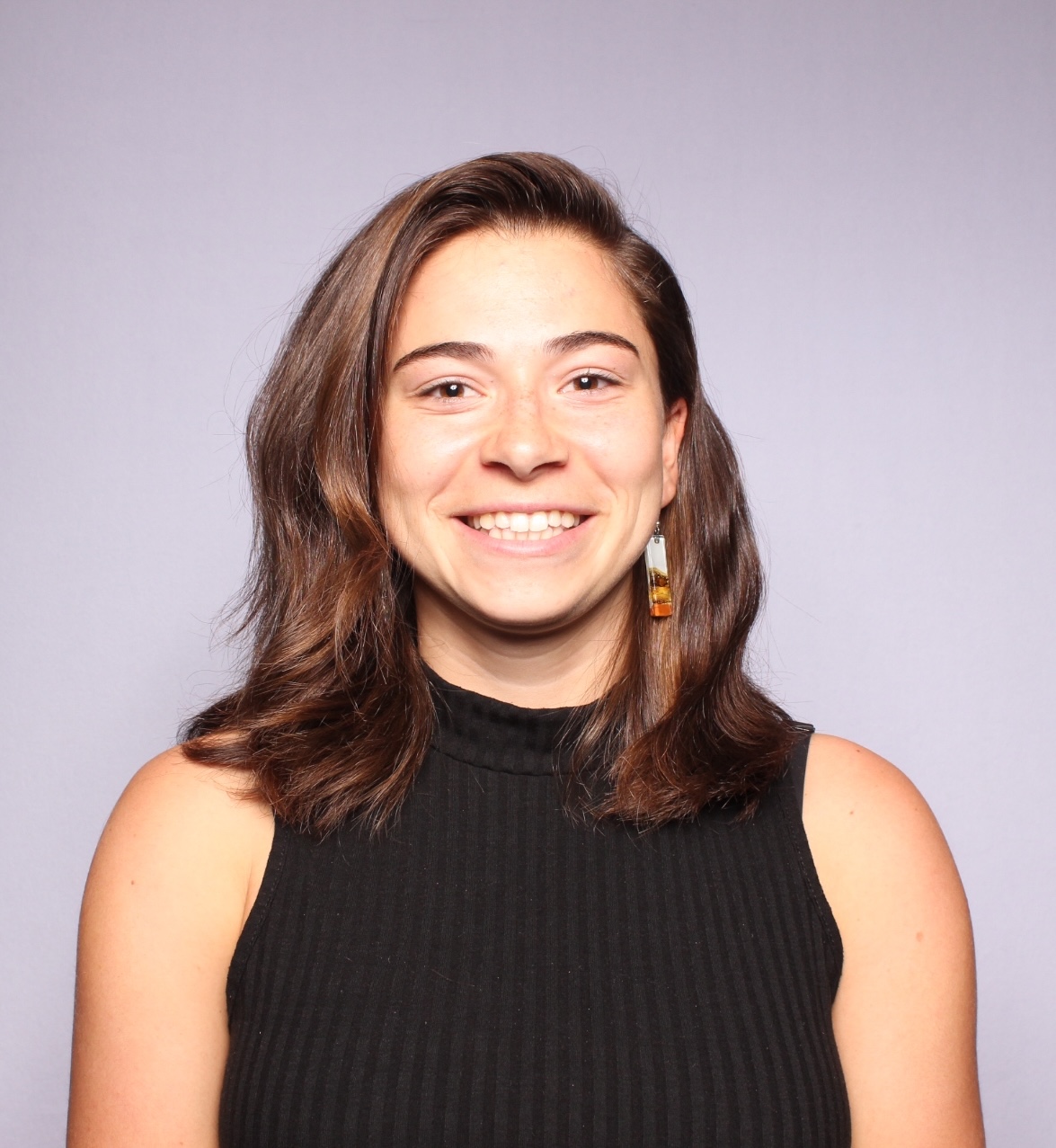}}]{Lilianna Houston} is a PhD candidate in the University of Denver's Physics and Astronomy Department. She works in Dr. Kingshuk Ghosh's lab on physics and machine learning models to better understand disordered proteins. She received a B.S. in Astrophysics from Tufts University in 2021.
\end{IEEEbiography}

\begin{IEEEbiography}[{\includegraphics[width=1in,height=1.25in,clip,keepaspectratio]{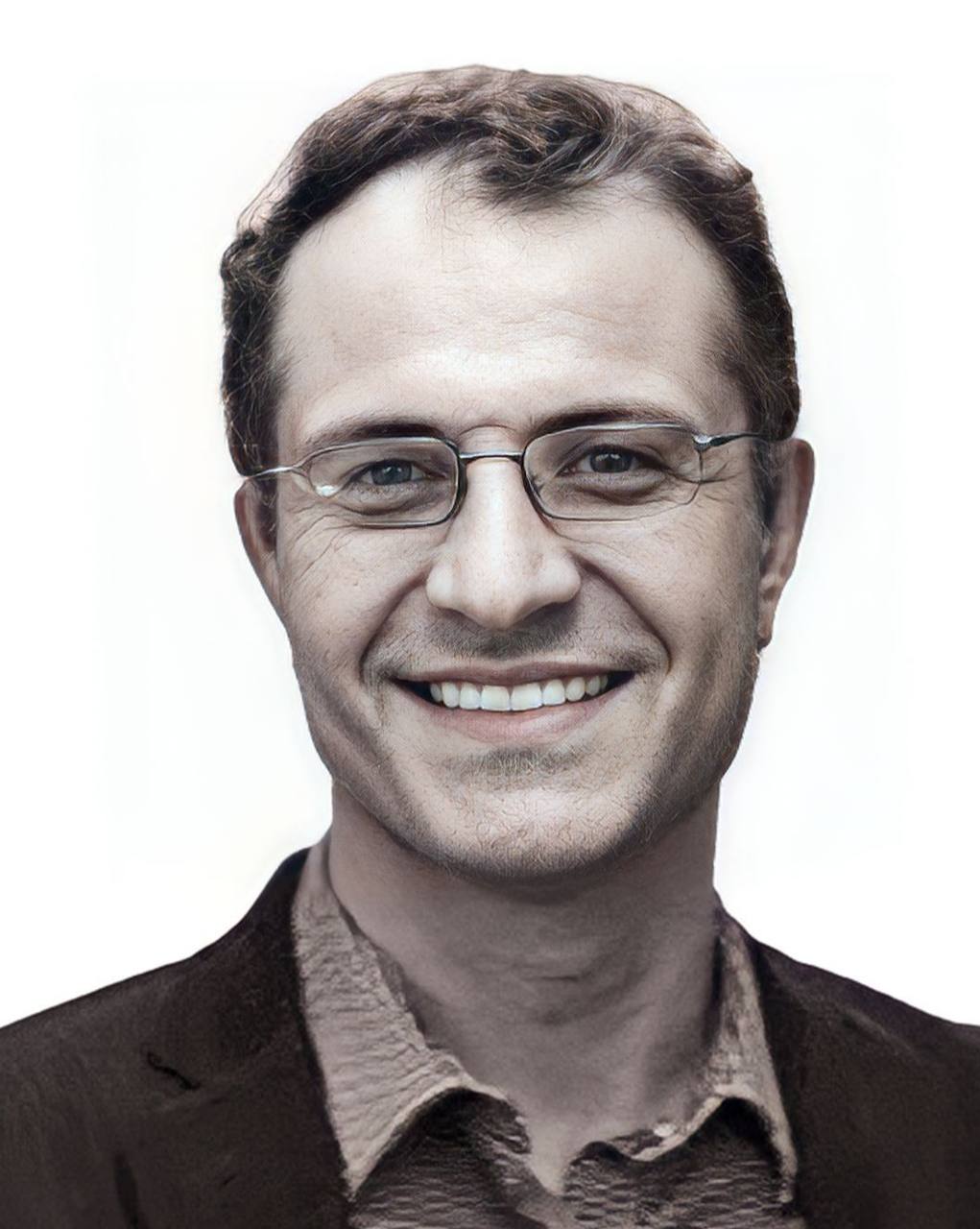}}]{Mohammad H. Mahoor} received an MS in Biomedical Engineering from Sharif University of Technology in 1998 and a Ph.D. in Electrical and Computer Engineering from the University of Miami in 2007. 
Currently a professor of Computer Science at the University of Denver, his research focuses on computer vision, deep machine learning, affective computing, and human-robot interaction, particularly with humanoid robots for children with autism and older adults with depression and dementia.
\end{IEEEbiography}

\vfill

\end{document}

%% file: Figures/S3_1_Model_Structure.tex
\begin{figure}[ht]
\vspace{-0.3cm}
\begin{center}
\includegraphics[width=8cm,height=2.8cm]{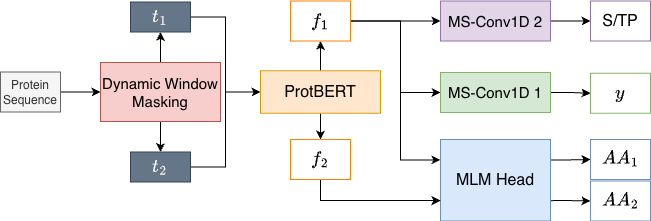}
\end{center}
\vspace{-0.3cm}
\caption{
Overall architecture of MT-ProtBERT. 
The input sequence is augmented twice via Dynamic Window Masking to generate $t_{1}$ and $t_{2}$, which are encoded by ProtBERT into embeddings $f_{1}$ and $f_{2}$. 
Three heads are applied: two MS-Conv1D classifiers and one MLM head. 
MS-Conv1D$_1$ and MS-Conv1D$_2$ use $f_{1}$ to predict the protein label ($y$, Task 1/2) and the S/T-P feature, respectively, while the MLM head reconstructs masked amino acids $AA_{1}$ and $AA_{2}$ from $f_{1}$ and $f_{2}$. 
These joint objectives enable MT-ProtBERT to learn sequence semantics and physical priors under data scarcity.
}
\label{fig: S3_1_MT-ProtBERT}
\vspace{-0.3cm}
\end{figure}

%% file: Figures/S3_4_MS-Conv1D.tex
\begin{figure}[ht]
\vspace{-0.3cm}
\begin{center}
\includegraphics[width=4.2cm,height=4.7cm]{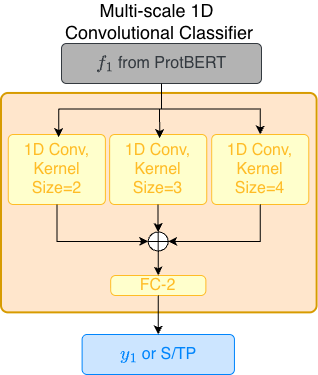}
\end{center}
\vspace{-0.3cm}
\caption{
Architecture of the Multi-scale 1D Convolutional Classifier (MS-Conv1D). 
Given an input embedding $f_{1}$ of size [batch size, embedding length = 768], three parallel 1D convolutional layers with kernel sizes of 2, 3, and 4 are used to extract local and global context features at multiple scales. 
Each branch outputs a feature of size [batch size, 128], which are concatenated into [batch size, 128*3 = 384] and passed through a fully connected layer to produce final prediction scores $y$ or S/TP of size [batch size, 2]. 
This design enhances feature richness and improves robustness under limited data.
}
\label{fig: S3_4_ms_conv1d}
\vspace{-0.3cm}
\end{figure}

%% file: Tables/S3_6_Chemical_Equivalence_Category.tex
\begin{table}[ht]
\centering
\caption{Chemical Equivalence Category for 20 AAs.}
\resizebox{5cm}{!}{
\begin{tabular}{l|l}
\hline
Polar residues ($\mathbf{\mu}$) & S,T,N,Q,C,H \\
Hydrophobic residues (h) & I,L,M,V \\
Basic residues (+) & R,K \\
Acidic residues (–) & E,D \\
Aromatic residues ($\mathbf{\pi}$) & F,W,Y \\
Alanine (A) & A \\
Proline (P) & P \\
Glycine (G) & G \\
\hline
\end{tabular}}
\label{tab: S3_6_chem_eqvl_ctgry}
\end{table}

%% file: Tables/S3_6_symbols_for_Protein_Aware_Loss.tex
\begin{table}[ht]
\centering
\caption{Symbols appeared in the Protein-aware loss function for MLM.}
\resizebox{8cm}{!}{
\begin{tabular}{c|l}
\hline
 Symbol & Description\\
\hline
 $x_{i}$ & Grounth truth AA at position i \\
 \(\tilde{x}\) & Masked sequence \\
 $\mathcal{M}$ & List of masked AA \\
 AA & AA dictionary with the length of 20 \\
 aa & Predicted AA \\
 \multirow{2}{*}{$G(x_{i})$} & \multirow{2}{6cm}{Chemical Equivalence Category for $x_{i}$. E.g. G(D)=\{D,E\}} \\
 & \\
 $|G(x_{i})|$ & Number of AA in $x_{i}$ group\\
 \multirow{2}{*}{$\epsilon$} & \multirow{2}{6cm}{Ratio, an hyperparameter to assign a weight for soft correct case. $\epsilon\in (0,1)$} \\
 & \\
 $p_{\theta_{i}}(aa|\tilde{x})$ & Probability of predicting as aa at position i \\
 $q(aa|x_{i})$ & Soft accuracy factor \\
\hline
\end{tabular}}
\label{tab: S3_6_protein_loss_symbol}
\end{table}

%% file: Tables/S4_0_Sample_Display.tex
\begin{table}[ht!]
\centering
\caption{
Example samples from PPA and PELM datasets. 
Each row corresponds to a 19-length protein sequence with its central amino acid (AA) and label indicating phosphorylation (Active) or no-phosphorylation (Non-active). 
These examples illustrate the short-sequence and balanced-label characteristics of both datasets.
}
\resizebox{6.6cm}{!}{
\begin{tabular}{lccc}
\hline
Type & AA & Text & Label \\
\hline
PPA & S & GSSLNSPKASFNKSSRFFS & 1 \\
PPA & T & NINRNTNQQTDIDHYQYVS & 0 \\
PPA & Y & FRQYGLWDRYADLYPQNDL & 0 \\
PELM & S & SVSHKKHSSSSEKTLHSKY & 0 \\
PELM & T & EGLPSEAALTPRPEGKVPS & 1 \\
PELM & Y & AKVDTDGNGYISFNELNDL & 1 \\
\hline
\end{tabular}}
\label{tab: S4_0_data_smp}
\end{table}

%% file: Tables/S4_1_Data_Distribution.tex
\begin{table}[ht!]
\centering
\caption{
Statistical summary of PPA and PELM datasets. 
AA denotes the central amino acid type in each sequence. 
While both datasets are label-balanced overall, subsets with AA=T or Y contain far fewer samples, 
revealing significant data scarcity that challenges model training.
}
\resizebox{4.5cm}{!}{
\begin{tabular}{llcc}
\hline
 & & \multicolumn{2}{c}{Label} \\ 
\cmidrule(lr){3-4}
Dataset & AA & Positive & Negative \\
\hline
\multirow{3}{*}{PPA} & S & 4121 & 4121 \\
 & T & 1361 & 1361 \\
 & Y & 578 & 578 \\
\hline
\multirow{3}{*}{PELM} & S & 15582 & 15582 \\
 & T & 4591 & 4591 \\
 & Y & 1768 & 1768 \\
\hline
\end{tabular}}
\label{tab: S4_1_dataset_stats}
\end{table}

%% file: Tables/S4_2_Tsk1_Exp1_Results.tex
\begin{table*}[ht]
\centering
\caption{
Comparison of phosphorylation prediction results between MT-ProtBERT and existing methods on PPA and PELM datasets. 
Our model achieves superior performance across most subsets, 
particularly on data-scarce groups such as PPA (AA=Y and AA = T), 
demonstrating its robustness and effectiveness under limited data.
}
\resizebox{18cm}{!}{
\begin{tabular}{llcccccc|cccccc}
\hline
\multirow{2}{0.5cm}{AA} & \multirow{2}{1cm}{Models} & \multicolumn{6}{c}{PPA} & \multicolumn{6}{c}{PELM} \\
\cline{3-14}
 & & Accuracy (\%) & F1 Score (\%) & AUPRC (\%) & Sensitivity (\%) & Specificity (\%) & MCC (\%) & Accuracy (\%) & F1 Score (\%) & AUPRC (\%) & Sensitivity (\%) & Specificity (\%) & MCC (\%) \\
\hline
 \multirow{4}{0.5cm}{S} & PARROT~\cite{S4_1_3_Parrot_2021} & 69.90 & 62.50 & 77.30 & 73.4 & 67.50 & 40.20 & 75.90$\pm$0.3 & 75.90$\pm$0.9 & & 75.80$\pm$0.4 & 76.10$\pm$0.6 & 51.90$\pm$0.6 \\
 & MusiteDeep~\cite{S4_4_14_MuDeep_2017} & 58.20 & 94.60 & 21.70 & 54.70 & 69.30 & 23.90 & 60.10$\pm$0.2 & 71.10$\pm$0.2 & & 98.35$\pm$0.07 & 21.90$\pm$0.2 & 31.40$\pm$0.3 \\
 & MT-ProtBERT & 
 72.74$\pm$1.95 &
 71.79$\pm$2.17 &
 67.55$\pm$1.52 &
 76.78$\pm$3.94 &
 68.67$\pm$3.65 &
 45.68$\pm$3.80 
 & 77.45$\pm$0.93 & 75.74$\pm$1.39 & 79.10$\pm$2.21 & 54.91$\pm$1.86 & 79.10$\pm$2.21 & 54.91$\pm$1.86 \\
 & & & & & & & & & & & & & \\
 
 \cline{3-14}
 \multirow{4}{0.5cm}{T} & PARROT~\cite{S4_1_3_Parrot_2021} & 59.80 & 39.90 & 79.70 & 66.3 & 49.80 & 21.00 & 72.90$\pm$0.6 & 72.60$\pm$0.7 & & 72.00$\pm$1.1 & 73.70$\pm$1.4 & 46.00$\pm$1.0 \\
 & MusiteDeep~\cite{S4_4_14_MuDeep_2017} & 59.70 & 56.40 & 63.10 & 60.40 & 58.30 & 19.50 & 73.20$\pm$0.2 & 76.30$\pm$0.3 & & 86.40$\pm$0.4 & 60.00$\pm$0.6 & 48.10$\pm$0.4 \\
 & MT-ProtBERT &
 66.40$\pm$1.78 & 
 67.31$\pm$1.96 & 
 59.67$\pm$2.45 & 
 62.03$\pm$3.85 &
 71.18$\pm$4.99 & 
 33.34$\pm$3.92
 & 77.67 $\pm$ 0.92 & 77.56 $\pm$ 1.38 & 71.95 $\pm$ 1.11 & 78.57 $\pm$ 4.01 & 76.70 $\pm$ 3.79 & 55.42 $\pm$ 1.84 \\
 & & & & & & & & & & & & & \\
 \cline{3-14}
 \multirow{4}{0.5cm}{Y} & 
 PARROT~\cite{S4_1_3_Parrot_2021} & 56.00 & 45.50 & 66.40 & 57.5 & 50.80 & 12.20 & 63.00$\pm$0.7 & 63.70$\pm$0.7 & & 65.20$\pm$1.0 & 61.00$\pm$2.0 & 26.00$\pm$1.0 \\
 & MusiteDeep~\cite{S4_4_14_MuDeep_2017} & 54.50 & 57.40 & 51.60 & 54.20 & 55.80 & 9.00 & 66.60$\pm$0.5 & 71.40$\pm$0.6 & & 83.50$\pm$0.6 & 49.70$\pm$0.9 & 35.30$\pm$0.7 \\
 & MT-ProtBERT &
 65.57$\pm$1.58 & 
 68.44$\pm$3.49 &
 61.79$\pm$5.72 &
 58.35$\pm$8.55 &
 72.55$\pm$8.28 &
 31.42$\pm$2.96
 & 68.05 $\pm$ 1.03 & 65.35 $\pm$ 5.01 & 60.26 $\pm$ 2.01 & 71.26 $\pm$ 8.14 & 64.07 $\pm$ 10.26 & 35.92 $\pm$ 2.33 \\
 & & & & & & & & & & & & & \\
\hline
\end{tabular}}
\label{tab: Tsk1_Exp1_results}
\end{table*}

%% file: Tables/S4_3_Tsk2_Exp2_4_Results.tex
\begin{table}[ht]
\centering
\caption{
Results of protein compaction prediction from different experiments and comparison between MT-ProtBERT and PARROT. 
MT-ProtBERT consistently achieves the highest mean Accuracy and MCC.
}
\resizebox{8.5cm}{!}{
\begin{tabular}{lcccccc}
\hline
 \multirow{3}{1cm}{Setting} & \multicolumn{6}{c}{Experiment 2} \\
\cline{2-7}
 & Accuracy (\%) & F1 Score (\%) & AUPRC (\%) & Sensitivity (\%) & Specificity (\%) & MCC (\%) \\
\hline
 PARROT & 67.25 $\pm$ 1.97 & 64.75 $\pm$ 4.46 & 61.86 $\pm$ 1.84 & 61.01 $\pm$ 7.28 & 73.39 $\pm$ 3.89 & 34.83 $\pm$ 3.58 \\
Optimized PARROT & 67.54 $\pm$ 3.38 & 63.27 $\pm$ 5.99 & 62.99 $\pm$ 3.24 & 57.48 $\pm$ 11.17 & 77.33 $\pm$ 11.73 & 36.85 $\pm$ 7.60 \\
 MT-ProtBERT & 
 76.18 $\pm$ 3.01 &
 74.52 $\pm$ 4.12 &
 72.09 $\pm$ 7.15 &
 82.79 $\pm$ 8.89 &
 68.72 $\pm$ 5.63 &
 52.12 $\pm$ 5.85 \\
\hline
 \multirow{3}{1cm}{Setting} & \multicolumn{6}{c}{Experiment 3 and 4} \\
\cline{2-7}
 & Accuracy (\%) & F1 Score (\%) & AUPRC (\%) & Sensitivity (\%) & Specificity (\%) & MCC (\%) \\
\hline
 PARROT & 62.26 $\pm$ 4.30 &
60.04 $\pm$ 6.21 &
58.16 $\pm$ 3.23 &
57.15 $\pm$ 8.66 &
67.41 $\pm$ 5.28 &
24.78 $\pm$ 8.47 \\
Optimized PARROT &
64.15 $\pm$ 5.20 &
65.54 $\pm$ 3.76 &
59.63 $\pm$ 4.26 &
67.78 $\pm$ 6.93 &
60.63 $\pm$ 12.50 &
28.85 $\pm$ 10.11 \\
MT-ProtBERT (Experiment 3) & 
74.34 $\pm$ 2.15 & 
73.13 $\pm$ 2.38 &	
66.19 $\pm$ 5.68 &
74.90 $\pm$ 5.89 &
74.07 $\pm$ 3.69 &
48.76 $\pm$ 4.27 \\

MT-ProtBERT (Experiment 4) & 
77.74 $\pm$ 3.10 & 
76.62 $\pm$ 4.95 & 
71.91 $\pm$ 1.88 & 
79.53 $\pm$ 13.44 & 
75.55 $\pm$ 14.22 & 
56.79 $\pm$ 5.57 \\
\hline
\end{tabular}}
\label{tab: Tsk2_Exp2_4_results}
\end{table}

%% file: Tables/S4_4_MLM_STP_SCD_Results.tex
\begin{table}[ht]
\centering
\caption{
Results of the auxiliary tasks besides main Task (1 or 2): Masked Language Modeling (MLM) and the binary task (S/T-P or SCD classification) Prediction. 
MT-ProtBERT achieves stable soft Top-1 and Top-5 accuracies in MLM, near-perfect performance in S/T-P task, and stable SCD Prediction, 
confirming that the model effectively learns sequence structure and biochemical properties while enhancing main-task performance.}
\resizebox{8.6cm}{!}{
\begin{tabular}{llccccc}
\hline
\multirow{2}{1cm}{Dataset} & \multirow{2}{0.5cm}{AA} & \multicolumn{4}{c}{MLM} & S/TP Prediction \\
 \cline{3-7}
 & & Top-1 Accuracy (\%) & Top-5 Accuracy (\%) & PLL & MRR (\%) & Accuracy (\%) \\
 \hline
 \multirow{3}{1cm}{Task 1 \\ PPA} & S & 
 49.69$\pm$3.51 &
 75.70$\pm$3.26 & 
 -2.27$\pm$0.06 &
 55.67$\pm$2.98 &
 100.00$\pm$0.00 \\
 & T & 
 43.19$\pm$3.69 &
 68.36$\pm$4.36 &
 -2.40$\pm$0.07 &
 48.86$\pm$3.87 &
 100.00$\pm$0.00 \\
 & Y & 
 40.49$\pm$5.22 &
 65.06$\pm$6.35 &
 -2.53$\pm$0.13 &
 45.24$\pm$6.12 &
 100.00$\pm$0.00 \\
 \hline
 \multirow{3}{1cm}{Task 1 \\ PELM} & S & 
 49.74$\pm$1.32 & 
 76.76$\pm$0.98 & 
 -2.19$\pm$0.02 & 
 57.01$\pm$0.96 &
 100.00$\pm$0.00 \\
 & T & 
 52.51$\pm$1.18 &
 78.22$\pm$0.92 &
 -2.20$\pm$0.03 &
 58.42$\pm$0.89 &
 100.00$\pm$0.00 \\
 & Y & 
 46.58$\pm$2.61 & 
 74.45$\pm$2.55 &
 -2.36$\pm$0.04 & 
 54.78$\pm$2.40 &
 100.00$\pm$0.00 \\
 \hline
 \hline
 \multicolumn{2}{l}{Task 2, Experiment 2} & 50.45$\pm$4.68 & 70.43$\pm$4.34 & -2.16$\pm$0.14 & 55.04$\pm$6.85 & 100.00$\pm$0.00 \\
 \hline
 \multicolumn{2}{l}{Task 2, Experiment 3} & 28.86$\pm$3.80 & 50.57$\pm$3.25 & -2.76$\pm$0.09 & 36.65$\pm$4.58 & 100.00$\pm$0.00 \\
\hline
\multirow{2}{1cm}{Dataset} & \multirow{2}{0.5cm}{AA} & \multicolumn{4}{c}{MLM} & SCD Prediction \\
 \cline{3-7}
 & & Top-1 Accuracy (\%) & Top-5 Accuracy (\%) & PLL & MRR (\%) & Accuracy (\%) \\
 \hline
 \multicolumn{2}{l}{Task 2, Experiment 4} & 28.66$\pm$1.73 & 48.54$\pm$2.40 & -2.81$\pm$0.07 & 36.57$\pm$2.12 & 72.07$\pm$4.09 \\
\hline
\end{tabular}}
\label{tab: mlm_stp_scd_results}
\end{table}

%% file: Figures/S4_7_Tsk1_Exp1_Abl_plot.tex
\begin{figure*}[ht]
\begin{center}
\includegraphics[width=18cm,height=11cm]{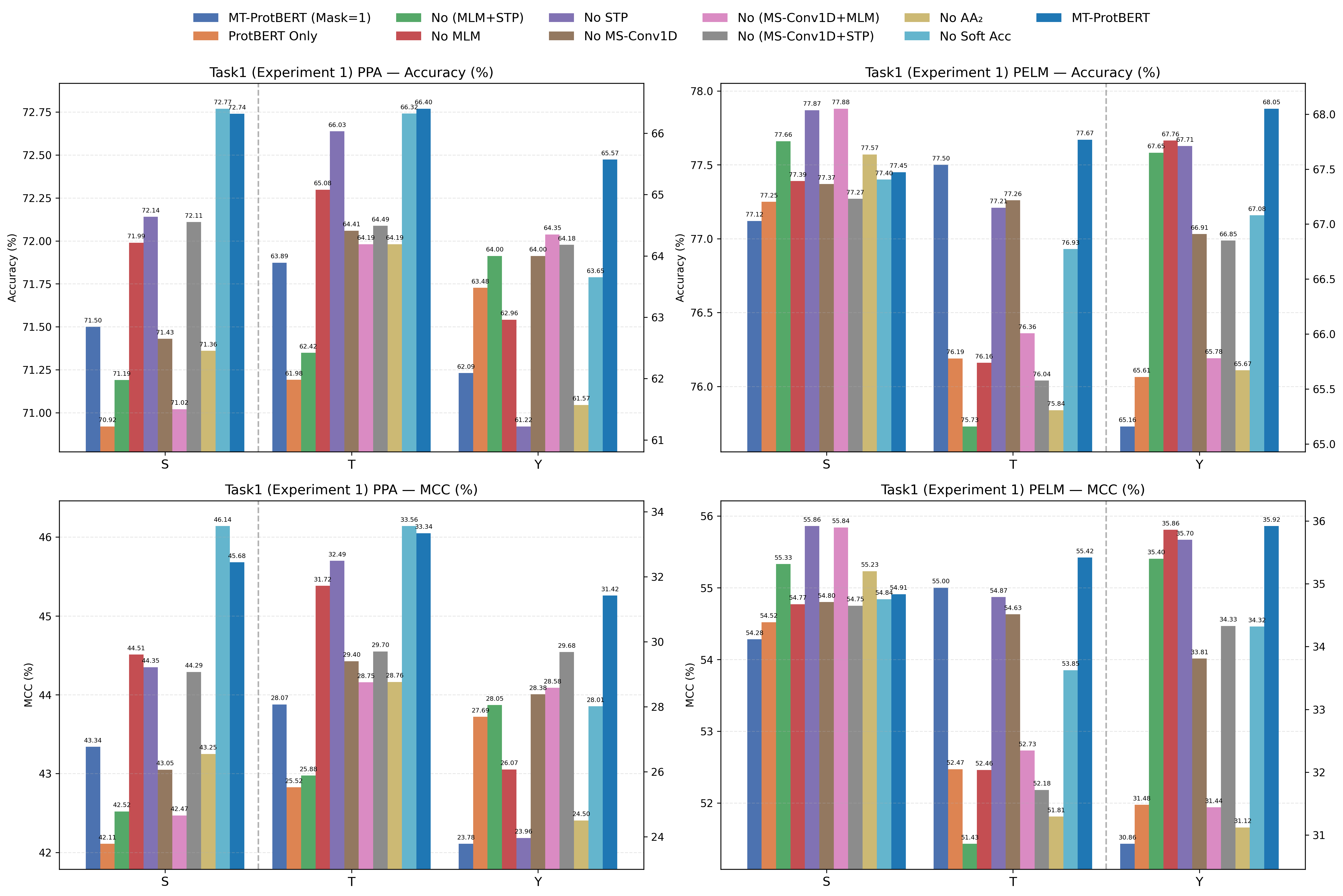}
\end{center}
\vspace{-0.3cm}
\caption{
Ablation study results on Task 1 in PPA and PELM datasets (Experiment 1). 
Each bar chart compares model variants by Accuracy and MCC, 
demonstrating the effectiveness of the proposed Dynamic Window Masking, MS-Conv1D, and multi-task learning. 
MT-ProtBERT consistently achieves the best balance between accuracy and correlation, confirming the contribution of each module.
}
\label{fig: S4_7_tsk1_exp1_abl_plot}
\vspace{-0.3cm}
\end{figure*}

%% file: Figures/S4_7_Tsk2_Exp2_Abl_plot.tex
\begin{figure*}[ht]
\begin{center}
\includegraphics[width=15cm,height=5cm]{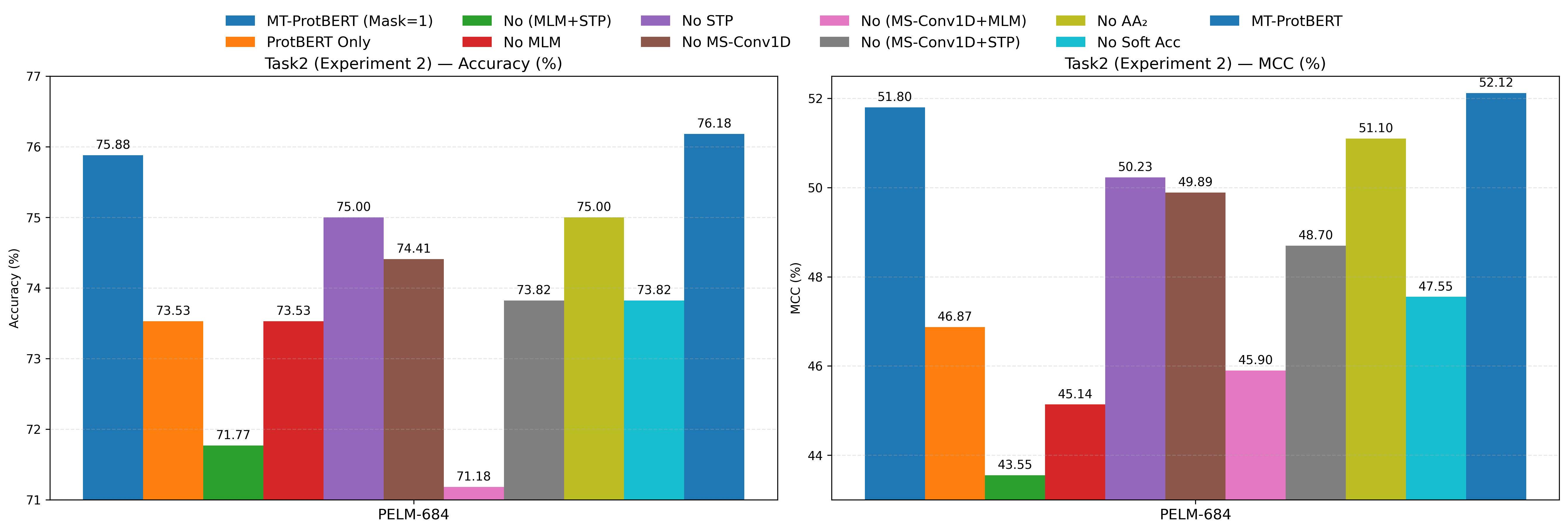}
\end{center}
\vspace{-0.3cm}
\caption{
Ablation study results for Experiment 2. Each bar chart compares model variants by Accuracy and MCC, demonstrating the effectiveness of the proposed Dynamic Window Masking, MS-Conv1D, and multi-task learning. MT-ProtBERT consistently achieves the best balance between accuracy and correlation, confirming the contribution of each module.
}
\label{fig: S4_7_tsk2_exp2_abl_plot}
\end{figure*}

%% file: Figures/S4_7_Tsk2_Exp3_Abl_plot.tex
\begin{figure*}[ht]
\begin{center}
\includegraphics[width=15cm,height=5cm]{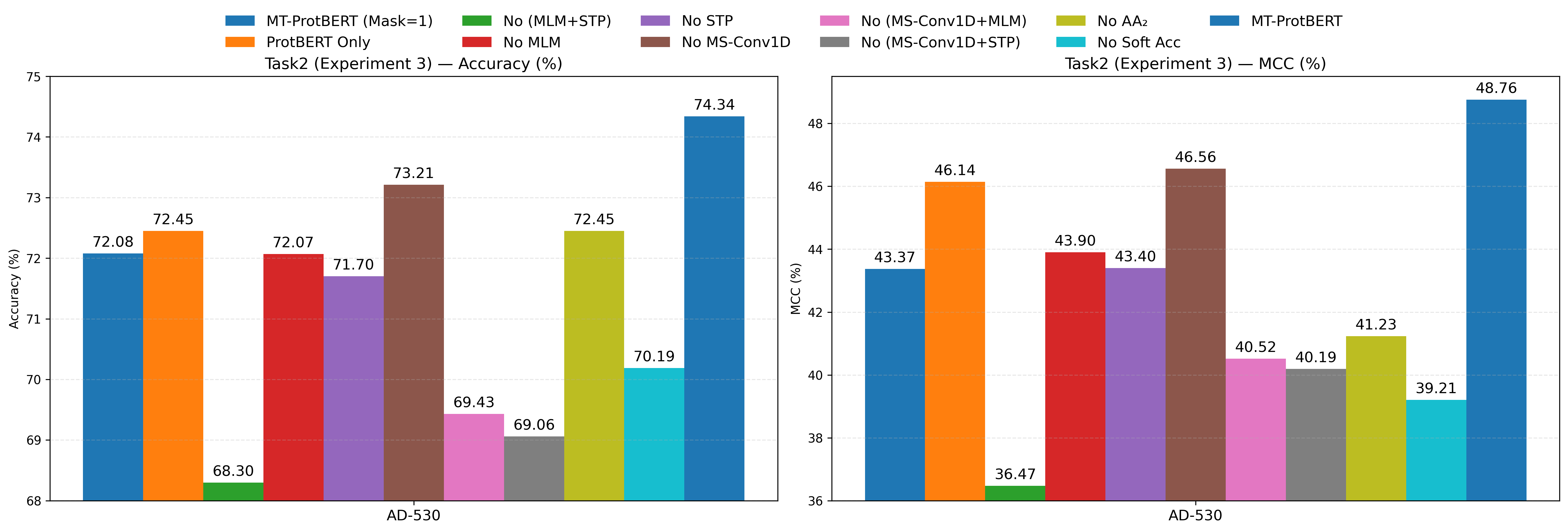}
\end{center}
\vspace{-0.3cm}
\caption{
Ablation study results for Experiment 3. Each bar chart compares model variants by Accuracy and MCC, demonstrating the effectiveness of the proposed Dynamic Window Masking, MS-Conv1D, and multi-task learning. MT-ProtBERT consistently achieves the best balance between accuracy and correlation, confirming the contribution of each module.
}
\label{fig: S4_7_tsk2_exp3_abl_plot}
\end{figure*}

%% file: Figures/S4_7_Tsk2_Exp4_Abl_plot.tex
\begin{figure*}[ht]
\vspace{-0.3cm}
\begin{center}
\includegraphics[width=15cm,height=5cm]{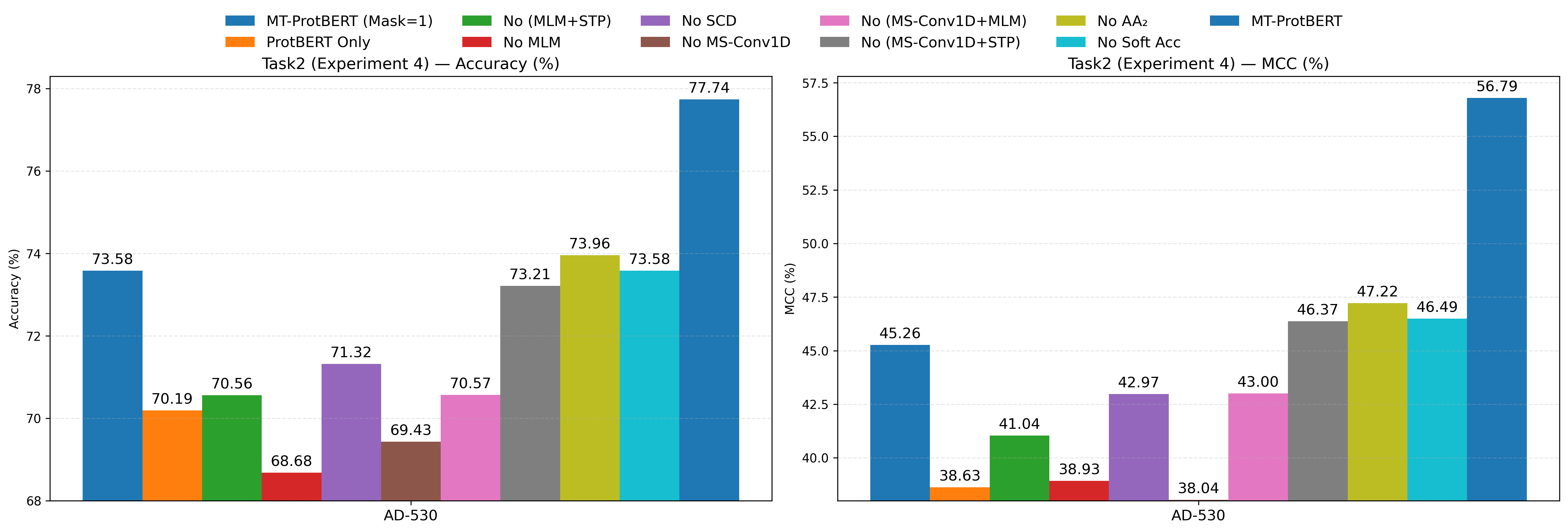}
\end{center}
\vspace{-0.3cm}
\caption{
Ablation study results for Experiment 4 (Task 2 with binary SCD prediction for the third dataset). 
Each bar chart compares model variants by Accuracy and MCC, 
demonstrating the effectiveness of the proposed Dynamic Window Masking, MS-Conv1D, and multi-task learning. 
MT-ProtBERT consistently achieves the best balance between accuracy and correlation, confirming the contribution of each module.}
\label{fig: S4_7_tsk2_exp4_abl_plot}
\vspace{-0.3cm}
\end{figure*}

%% file: Tables/SApp_Ablt_Tsk1_Exp1.tex
\begin{table*}[ht]
\centering
\caption{
Detailed ablation study results comparing model variants on PPA and PELM datasets. 
MT-ProtBERT consistently achieves the highest mean Accuracy and MCC across all subsets, 
validating the necessity of each proposed module. 
\emph{MT-ProtBERT (Mask=1)} indicates single-position Dynamic Window Masking; 
\emph{ProtBERT Only} removes all proposed modules except the backbone and a fully connected output layer; 
\emph{No (MLM+STP)} disables multi-task learning; 
\emph{No STP} and \emph{No MLM} individually remove each auxiliary task;
\emph{No $AA_{2}$} removes the second MLM task;
\emph{No Soft Acc} switches soft Top-1 accuracy to hard one and deactivates Protein - aware Loss Function but use standard cross entropy loss;
\emph{No MS-Conv1D} replaces the multi-scale classifier with a single fully connected layer; 
and combinations such as \emph{No (MS-Conv1D+MLM)} or \emph{No (MS-Conv1D+STP)} disable both modules simultaneously. 
Results confirm that Dynamic Window Masking, MS-Conv1D, and multi-task learning jointly drive the performance gain.
}
\resizebox{18cm}{!}{
\begin{tabular}{llcccccc|cccccc}
\hline
\multirow{2}{0.5cm}{AA} & \multirow{2}{1cm}{Setting} & \multicolumn{6}{c}{PPA} & \multicolumn{6}{c}{PELM} \\
\cline{3-14}
 & & Accuracy (\%) & F1 Score (\%) & AUPRC (\%) & Sensitivity (\%) & Specificity (\%) & MCC (\%) & Accuracy (\%) & F1 Score (\%) & AUPRC (\%) & Sensitivity (\%) & Specificity (\%) & MCC (\%) \\
\hline
 \multirow{8}{0.5cm}{S} 
& MT-ProtBERT (Mask=1)
& 71.50 $\pm$ 0.63 & 73.97 $\pm$ 7.89 & 68.81 $\pm$ 8.45 & 43.34 $\pm$ 1.14 & 68.81 $\pm$ 8.45 & 43.34 $\pm$ 1.14 
& 77.12 $\pm$ 0.77 & 77.65 $\pm$ 2.24 & 76.61 $\pm$ 1.18 & 54.28 $\pm$ 1.52 & 76.61 $\pm$ 1.18 & 54.28 $\pm$ 1.52 \\

& ProtBERT Only
& 70.92 $\pm$ 1.10 & 76.57 $\pm$ 5.51 & 65.02 $\pm$ 5.82 & 42.11 $\pm$ 2.38 & 65.02 $\pm$ 5.82 & 42.11 $\pm$ 2.38
& 77.25 $\pm$ 0.72 & 78.11 $\pm$ 1.90 & 76.36 $\pm$ 2.48 & 54.52 $\pm$ 1.42 & 76.36 $\pm$ 2.48 & 54.52 $\pm$ 1.42 \\

& No (MLM+STP)
& 71.19 $\pm$ 1.23 & 77.98 $\pm$ 2.52 & 64.03 $\pm$ 2.82 & 42.52 $\pm$ 2.13 & 64.03 $\pm$ 2.82 & 42.52 $\pm$ 2.13
& 77.66 $\pm$ 0.13 & 76.35 $\pm$ 1.87 & 78.93 $\pm$ 1.81 & 55.33 $\pm$ 0.27 & 78.93 $\pm$ 1.81 & 55.33 $\pm$ 0.27 \\

& No MLM
& 71.99 $\pm$ 1.60 & 77.59 $\pm$ 6.05 & 66.51 $\pm$ 3.63 & 44.51 $\pm$ 3.92 & 66.51 $\pm$ 3.63 & 44.51 $\pm$ 3.92
& 77.39 $\pm$ 0.78 & 77.96 $\pm$ 2.27 & 76.75 $\pm$ 3.10 & 54.77 $\pm$ 1.48 & 76.75 $\pm$ 3.10 & 54.77 $\pm$ 1.48 \\

& No STP
& 72.14 $\pm$ 1.02 & 74.11 $\pm$ 3.59 & 70.10 $\pm$ 5.02 & 44.35 $\pm$ 2.13 & 70.10 $\pm$ 5.02 & 44.35 $\pm$ 2.13
& 77.87 $\pm$ 0.67 & 78.14 $\pm$ 3.53 & 77.56 $\pm$ 4.49 & 55.86 $\pm$ 1.39 & 77.56 $\pm$ 4.49 & 55.86 $\pm$ 1.39 \\

& No MS-Conv1D
& 71.43 $\pm$ 1.57 & 77.10 $\pm$ 2.82 & 65.57 $\pm$ 4.23 & 43.05 $\pm$ 3.06 & 65.57 $\pm$ 4.23 & 43.05 $\pm$ 3.06
& 77.37 $\pm$ 0.62 & 77.51 $\pm$ 2.72 & 77.20 $\pm$ 3.44 & 54.80 $\pm$ 1.28 & 77.20 $\pm$ 3.44 & 54.80 $\pm$ 1.28 \\

& No (MS-Conv1D+MLM)
& 71.02 $\pm$ 1.39 & 68.12 $\pm$ 4.00 & 65.49 $\pm$ 2.91 & 79.01 $\pm$ 4.41 & 62.70 $\pm$ 6.62 & 42.47 $\pm$ 2.89
& 77.88 $\pm$ 0.50 & 78.08 $\pm$ 0.81 & 71.08 $\pm$ 0.94 & 76.13 $\pm$ 2.38 & 79.64 $\pm$ 2.06 & 55.84 $\pm$ 0.90 \\

& No (MS-Conv1D+STP)
& 72.11 $\pm$ 1.23 & 76.27 $\pm$ 4.56 & 67.79 $\pm$ 2.77 & 44.29 $\pm$ 2.46 & 67.79 $\pm$ 2.77 & 44.29 $\pm$ 2.46
& 77.27 $\pm$ 1.12 & 77.04 $\pm$ 1.40 & 71.50 $\pm$ 1.24 & 78.53 $\pm$ 4.41 & 76.06 $\pm$ 4.50 & 54.75 $\pm$ 2.16 \\

& No $AA_{2}$
& 71.36 $\pm$ 0.98 & 76.51 $\pm$ 7.96 & 66.06 $\pm$ 7.76 & 43.25 $\pm$ 2.09 & 66.06 $\pm$ 7.76 & 43.25 $\pm$ 2.09
& 77.57 $\pm$ 0.64 & 75.77 $\pm$ 2.71 & 79.32 $\pm$ 3.73 & 55.23 $\pm$ 1.36 & 79.32 $\pm$ 3.73 & 55.23 $\pm$ 1.36 \\

& No Soft Acc
& 72.77 $\pm$ 1.04 & 71.34 $\pm$ 2.30 & 67.98 $\pm$ 2.63 & 78.49 $\pm$ 6.23 & 67.12 $\pm$ 5.43 & 46.14 $\pm$ 2.75
& 77.40 $\pm$ 0.57 & 77.97 $\pm$ 0.30 & 71.15 $\pm$ 0.77 & 75.30 $\pm$ 1.45 & 79.48 $\pm$ 0.81 & 54.84 $\pm$ 1.12 \\

& MT-ProtBERT &
 72.74$\pm$1.95 &
 71.79$\pm$2.17 &
 67.55$\pm$1.52 &
 76.78$\pm$3.94 &
 68.67$\pm$3.65 &
 45.68$\pm$3.80 & 
 77.45$\pm$0.93 & 75.74$\pm$1.39 & 79.10$\pm$2.21 & 54.91$\pm$1.86 & 79.10$\pm$2.21 & 54.91$\pm$1.86 \\
 \cline{2-14}
 \multirow{8}{0.5cm}{T} 
& MT-ProtBERT (Mask=1)
& 63.89 $\pm$ 3.66 & 66.57 $\pm$ 5.83 & 59.02 $\pm$ 3.97 & 55.02 $\pm$ 11.50 & 72.08 $\pm$ 11.07 & 28.07 $\pm$ 7.24 
& 77.50 $\pm$ 0.90 & 76.94 $\pm$ 1.11 & 70.57 $\pm$ 1.37 & 78.12 $\pm$ 2.43 & 76.84 $\pm$ 2.46 & 55.00 $\pm$ 1.81 \\

& ProtBERT Only
& 61.98 $\pm$ 4.30 & 68.42 $\pm$ 5.11 & 58.74 $\pm$ 3.48 & 42.74 $\pm$ 10.26 & 80.15 $\pm$ 12.02 & 25.52 $\pm$ 9.23
& 76.19 $\pm$ 1.15 & 76.34 $\pm$ 1.00 & 69.62 $\pm$ 1.72 & 75.01 $\pm$ 4.79 & 77.32 $\pm$ 3.21 & 52.47 $\pm$ 2.37 \\

& No (MLM+STP)
& 62.42 $\pm$ 2.46 & 61.81 $\pm$ 4.62 & 56.09 $\pm$ 2.64 & 60.52 $\pm$ 14.94 & 64.35 $\pm$ 13.87 & 25.88 $\pm$ 5.27
& 75.73 $\pm$ 0.76 & 76.34 $\pm$ 0.74 & 70.67 $\pm$ 1.12 & 75.38 $\pm$ 2.80 & 76.02 $\pm$ 1.98 & 51.43 $\pm$ 1.58 \\

& No MLM
& 65.08 $\pm$ 1.22 & 68.31 $\pm$ 5.37 & 59.36 $\pm$ 1.36 & 52.50 $\pm$ 14.38 & 76.81 $\pm$ 14.70 & 31.72 $\pm$ 3.62
& 76.16 $\pm$ 0.91 & 75.96 $\pm$ 1.23 & 70.12 $\pm$ 1.89 & 77.00 $\pm$ 3.91 & 75.37 $\pm$ 3.36 & 52.46 $\pm$ 1.88 \\

& No STP
& 66.03 $\pm$ 2.88 & 68.93 $\pm$ 4.93 & 61.00 $\pm$ 3.94 & 56.54 $\pm$ 11.57 & 74.71 $\pm$ 10.48 & 32.49 $\pm$ 5.46
& 77.21 $\pm$ 0.84 & 75.89 $\pm$ 1.33 & 71.24 $\pm$ 1.78 & 81.38 $\pm$ 6.06 & 72.98 $\pm$ 5.59 & 54.87 $\pm$ 1.81 \\

& No MS-Conv1D
& 64.41 $\pm$ 0.95 & 66.00 $\pm$ 6.03 & 58.26 $\pm$ 2.40 & 56.11 $\pm$ 14.35 & 71.64 $\pm$ 14.73 & 29.40 $\pm$ 2.70
& 77.26 $\pm$ 1.26 & 77.14 $\pm$ 1.31 & 72.23 $\pm$ 2.18 & 79.42 $\pm$ 2.28 & 75.21 $\pm$ 0.91 & 54.63 $\pm$ 2.58 \\

& No (MS-Conv1D+MLM)
& 64.19 $\pm$ 2.53 & 65.63 $\pm$ 5.14 & 58.55 $\pm$ 2.74 & 58.29 $\pm$ 9.23 & 69.76 $\pm$ 10.70 & 28.75 $\pm$ 5.29
& 76.36 $\pm$ 1.27 & 76.30 $\pm$ 1.36 & 70.59 $\pm$ 1.45 & 77.23 $\pm$ 1.55 & 75.49 $\pm$ 1.49 & 52.73 $\pm$ 2.53 \\

& No (MS-Conv1D+STP)
& 64.49 $\pm$ 1.53 & 68.71 $\pm$ 4.57 & 59.93 $\pm$ 2.01 & 51.01 $\pm$ 9.64 & 77.00 $\pm$ 11.07 & 29.70 $\pm$ 3.82
& 76.04 $\pm$ 1.19 & 75.99 $\pm$ 1.16 & 70.51 $\pm$ 0.78 & 77.19 $\pm$ 2.76 & 74.94 $\pm$ 3.03 & 52.18 $\pm$ 2.30 \\

& No $AA_{2}$
& 64.19 $\pm$ 2.35 & 66.77 $\pm$ 4.80 & 59.34 $\pm$ 1.87 & 55.83 $\pm$ 9.11 & 72.06 $\pm$ 11.00 & 28.76 $\pm$ 5.96
& 75.84 $\pm$ 1.30 & 74.76 $\pm$ 1.78 & 69.00 $\pm$ 1.57 & 78.13 $\pm$ 4.36 & 73.46 $\pm$ 4.62 & 51.81 $\pm$ 2.48 \\

& No Soft Acc
& 66.32 $\pm$ 2.53 & 63.97 $\pm$ 8.89 & 59.93 $\pm$ 3.53 & 67.72 $\pm$ 15.90 & 64.21 $\pm$ 16.82 & 33.56 $\pm$ 4.87
& 76.93 $\pm$ 1.69 & 76.05 $\pm$ 2.18 & 70.07 $\pm$ 2.36 & 78.51 $\pm$ 4.21 & 75.15 $\pm$ 4.18 & 53.85 $\pm$ 3.40 \\

& MT-ProtBERT & 
 66.40$\pm$1.78 & 
 67.31$\pm$1.96 & 
 59.67$\pm$2.45 & 
 62.03$\pm$3.85 &
 71.18$\pm$4.99 & 
 33.34$\pm$3.92 & 
 77.67 $\pm$ 0.92 & 77.56 $\pm$ 1.38 & 71.95 $\pm$ 1.11 & 78.57 $\pm$ 4.01 & 76.70 $\pm$ 3.79 & 55.42 $\pm$ 1.84 \\

 \cline{2-14}
 \multirow{8}{0.5cm}{Y} 
& MT-ProtBERT (Mask=1)
& 62.09 $\pm$ 0.99 & 65.66 $\pm$ 1.71 & 59.67 $\pm$ 1.25 & 54.40 $\pm$ 5.53 & 69.01 $\pm$ 5.11 & 23.78 $\pm$ 2.40 
& 65.16 $\pm$ 1.19 & 62.96 $\pm$ 8.53 & 60.26 $\pm$ 2.67 & 68.17 $\pm$ 13.28 & 61.47 $\pm$ 16.12 & 30.86 $\pm$ 2.26 \\

& ProtBERT Only
& 63.48 $\pm$ 1.07 & 64.72 $\pm$ 3.05 & 56.91 $\pm$ 1.59 & 57.55 $\pm$ 6.41 & 69.77 $\pm$ 7.40 & 27.69 $\pm$ 2.42
& 65.61 $\pm$ 2.60 & 64.82 $\pm$ 5.29 & 60.77 $\pm$ 3.36 & 67.56 $\pm$ 6.67 & 63.69 $\pm$ 9.37 & 31.48 $\pm$ 4.99 \\

& No (MLM+STP)
& 64.00 $\pm$ 3.17 & 67.69 $\pm$ 6.33 & 58.84 $\pm$ 4.15 & 50.13 $\pm$ 13.11 & 75.74 $\pm$ 12.55 & 28.05 $\pm$ 7.16
& 67.65 $\pm$ 1.29 & 66.14 $\pm$ 2.62 & 60.68 $\pm$ 2.46 & 69.63 $\pm$ 7.37 & 65.48 $\pm$ 6.49 & 35.40 $\pm$ 2.67 \\

& No MLM
& 62.96 $\pm$ 3.35 & 66.25 $\pm$ 6.30 & 57.18 $\pm$ 4.97 & 50.19 $\pm$ 8.35 & 74.58 $\pm$ 10.09 & 26.07 $\pm$ 6.23
& 67.76 $\pm$ 1.51 & 67.94 $\pm$ 2.20 & 63.28 $\pm$ 1.85 & 68.33 $\pm$ 10.04 & 67.12 $\pm$ 7.84 & 35.86 $\pm$ 2.99 \\

& No STP
& 61.22 $\pm$ 1.80 & 64.62 $\pm$ 3.95 & 55.03 $\pm$ 2.07 & 47.80 $\pm$ 16.10 & 74.36 $\pm$ 12.12 & 23.96 $\pm$ 3.67
& 67.71 $\pm$ 2.07 & 68.10 $\pm$ 3.08 & 62.33 $\pm$ 3.29 & 66.78 $\pm$ 3.13 & 68.88 $\pm$ 6.19 & 35.70 $\pm$ 4.01 \\

& No MS-Conv1D
& 64.00 $\pm$ 4.06 & 68.85 $\pm$ 3.37 & 60.58 $\pm$ 5.44 & 50.64 $\pm$ 13.90 & 76.44 $\pm$ 8.05 & 28.38 $\pm$ 6.06
& 66.91 $\pm$ 2.06 & 61.26 $\pm$ 4.31 & 60.22 $\pm$ 3.18 & 78.06 $\pm$ 4.51 & 54.57 $\pm$ 6.54 & 33.81 $\pm$ 3.41 \\

& No (MS-Conv1D+MLM)
& 64.35 $\pm$ 5.74 & 67.84 $\pm$ 6.69 & 61.11 $\pm$ 7.67 & 55.16 $\pm$ 13.14 & 72.58 $\pm$ 9.34 & 28.58 $\pm$ 10.81
& 65.78 $\pm$ 2.73 & 66.92 $\pm$ 3.67 & 61.98 $\pm$ 3.01 & 64.10 $\pm$ 5.46 & 67.22 $\pm$ 6.57 & 31.44 $\pm$ 5.66 \\

& No (MS-Conv1D+STP)
& 64.18 $\pm$ 5.48 & 64.95 $\pm$ 10.48 & 57.77 $\pm$ 4.61 & 54.87 $\pm$ 27.64 & 72.13 $\pm$ 22.81 & 29.68 $\pm$ 9.86
& 66.85 $\pm$ 1.78 & 63.11 $\pm$ 4.69 & 60.19 $\pm$ 2.20 & 73.37 $\pm$ 12.23 & 59.80 $\pm$ 12.19 & 34.33 $\pm$ 2.73 \\

& No $AA_{2}$ 
& 61.57 $\pm$	3.10 & 66.54 $\pm$ 3.59 & 56.96 $\pm$ 2.59 & 46.27 $\pm$	7.93 & 76.83 $\pm$ 8.42 & 24.50 $\pm$ 7.45
& 65.67 $\pm$ 2.48 & 61.46 $\pm$ 7.11 & 58.52 $\pm$ 3.93 & 71.64 $\pm$ 11.60 & 58.50 $\pm$ 13.11 & 31.12 $\pm$ 5.41 \\

& No Soft Acc
& 63.65 $\pm$ 2.71 & 68.44 $\pm$ 5.59 & 59.61 $\pm$ 3.51 & 48.73 $\pm$ 9.18 & 77.42 $\pm$ 11.74 & 28.01 $\pm$ 4.75
& 67.08 $\pm$ 1.01 & 64.21 $\pm$ 4.68 & 60.67 $\pm$ 2.42 & 71.82 $\pm$ 12.03 & 61.45 $\pm$ 11.29 & 34.32 $\pm$ 2.16 \\

& MT-ProtBERT &
 65.57$\pm$1.58 & 
 68.44$\pm$3.49 &
 61.79$\pm$5.72 &
 58.35$\pm$8.55 &
 72.55$\pm$8.28 &
 31.42$\pm$2.96 & 
 68.05 $\pm$ 1.03 & 65.35 $\pm$ 5.01 & 60.26 $\pm$ 2.01 & 71.26 $\pm$ 8.14 & 64.07 $\pm$ 10.26 & 35.92 $\pm$ 2.33 \\
\hline
\end{tabular}}
\label{tab: SApp_1_Abl_Tsk1_Exp1}
\end{table*}

%% file: Tables/SApp_Ablt_Tsk2_Exp2.tex
\begin{table}[ht]
\centering
\caption{
Detailed ablation study results comparing model variants for Experiment 2. 
MT-ProtBERT consistently achieves the highest mean Accuracy and MCC across all subsets, validating the necessity of each proposed module. 
Results confirm that Dynamic Window Masking, MS-Conv1D, and multi-task learning jointly drive the performance gain.
}
\resizebox{8.5cm}{!}{
\begin{tabular}{lcccccc}
\hline
 \multirow{2}{1cm}{Setting} & \multicolumn{6}{c}{Dataset 2 with 684 Samples} \\
\cline{2-7}
 & Accuracy (\%) & F1 Score (\%) & AUPRC (\%) & Sensitivity (\%) & Specificity (\%) & MCC (\%) \\
\hline
 MT-ProtBERT (Mask=1) & 75.88 $\pm$ 3.97 & 72.78 $\pm$ 7.59 & 78.98 $\pm$ 5.04 & 51.80 $\pm$ 8.90 & 78.98 $\pm$ 5.04 & 51.80 $\pm$ 8.90 \\
 ProtBERT Only & 73.53$\pm$3.45 & 80.30$\pm$4.99 & 65.92$\pm$12.04 & 46.87$\pm$7.70 & 65.92$\pm$12.04 & 46.87$\pm$7.70 \\
 No (MLM+STP) & 71.77$\pm$3.01 & 70.59$\pm$13.40 & 71.64$\pm$14.59 & 43.55$\pm$5.93 & 71.64$\pm$14.59 & 43.55$\pm$5.93 \\
 No MLM & 73.53$\pm$5.97 & 79.25$\pm$11.51 & 62.80$\pm$20.20 & 45.14$\pm$12.40 & 62.80$\pm$20.20 & 45.14$\pm$12.40 \\
 No STP &
 75.00$\pm$3.29 &
 80.54$\pm$6.47 &
 69.16$\pm$5.47 &
 50.23$\pm$6.68 &
 69.16$\pm$5.47 &
 50.23$\pm$6.68 \\
 No MS-Conv1D & 74.41$\pm$7.10 & 84.32$\pm$12.46 & 63.82$\pm$4.21 & 49.89$\pm$14.43 & 63.82$\pm$4.21 & 49.89$\pm$14.43 \\
 No (MS-Conv1D+MLM) & 71.18$\pm$3.97 & 80.47$\pm$11.31 & 63.99$\pm$16.27 & 45.90$\pm$4.67 & 63.99$\pm$16.27 & 45.90$\pm$4.67 \\
 No (MS-Conv1D+STP) & 73.82$\pm$8.01 & 79.33$\pm$6.44 & 69.34$\pm$10.12 & 48.70$\pm$14.59 & 69.34$\pm$10.12 & 48.70$\pm$14.59 \\
 No $AA_{2}$ & 75.00$\pm$1.80 & 82.30$\pm$7.45 & 67.67$\pm$10.05 & 51.10$\pm$3.12 & 67.67$\pm$10.05 & 51.10$\pm$3.12 \\
 No Soft Acc & 73.82$\pm$1.92 & 76.86$\pm$3.46 & 70.62$\pm$6.95 & 47.55$\pm$3.40 & 70.62$\pm$6.95 & 47.55$\pm$3.40 \\
 MT-ProtBERT & 
 76.18$\pm$3.01 &
 82.79$\pm$8.89 &
 68.72$\pm$5.63 &
 52.12$\pm$5.85 &
 68.72$\pm$5.63 &
 52.12$\pm$5.85 \\
\hline
\end{tabular}}
\label{tab: SApp_1_Abl_Tsk2_Exp2}
\end{table}

%% file: Tables/SApp_Ablt_Tsk2_Exp3.tex
\begin{table}[ht]
\centering
\caption{
Detailed ablation study results comparing model variants for Experiment 3. 
MT-ProtBERT consistently achieves the highest mean Accuracy and MCC across all subsets, validating the necessity of each proposed module. 
Results confirm that Dynamic Window Masking, MS-Conv1D, and multi-task learning jointly drive the performance gain.
}
\resizebox{8.5cm}{!}{
\begin{tabular}{lcccccc}
\hline
 \multirow{2}{1cm}{Setting} & \multicolumn{6}{c}{Dataset 3 with 530 Samples} \\
\cline{2-7}
 & Accuracy (\%) & F1 Score (\%) & AUPRC (\%) & Sensitivity (\%) & Specificity (\%) & MCC (\%) \\
\hline
 MT-ProtBERT (Mask=1) & 72.08 $\pm$ 3.87 & 72.17 $\pm$ 11.51 & 70.69 $\pm$ 7.74 & 43.37 $\pm$ 7.25 & 70.69 $\pm$ 7.74 & 43.37 $\pm$ 7.25 \\
 ProtBERT Only & 72.45$\pm$6.48 & 76.95$\pm$9.29 & 68.98$\pm$12.61 & 46.14$\pm$12.28 & 68.98$\pm$12.61 & 46.14$\pm$12.28 \\
 No (MLM+STP) & 68.30$\pm$4.70 & 64.96$\pm$10.65 & 71.32$\pm$4.14 & 36.47$\pm$8.94 & 71.32$\pm$4.14 & 36.47$\pm$8.94 \\
 No MLM &
 72.07 $\pm$ 3.38 & 
 70.70 $\pm$ 7.78 &
 72.63 $\pm$ 9.14 & 
 43.90 $\pm$ 6.96 &
 72.63 $\pm$ 9.14 & 
 43.90 $\pm$ 6.96 \\
 No STP & 71.70$\pm$4.42 & 76.40$\pm$8.06 & 66.39$\pm$8.93 & 43.40$\pm$9.37 & 66.39$\pm$8.93 & 43.40$\pm$9.37 \\
 No MS-Conv1D & 73.21$\pm$4.51 & 75.09$\pm$7.73 & 70.82$\pm$14.66 & 46.56$\pm$11.45 & 70.82$\pm$14.66 & 46.56$\pm$11.45 \\
 No (MS-Conv1D+MLM) & 69.43$\pm$6.46 & 58.84$\pm$16.64 & 80.47$\pm$6.85 & 40.52$\pm$12.18 & 80.47$\pm$6.85 & 40.52$\pm$12.18 \\
 No (MS-Conv1D+STP) & 69.06$\pm$7.96 & 63.69$\pm$27.90 & 73.14$\pm$16.83 & 40.19$\pm$13.02 & 73.14$\pm$16.83 & 40.19$\pm$13.02 \\
 No $AA_{2}$ & 72.45$\pm$5.10 & 82.59$\pm$17.08 & 53.95$\pm$30.24 & 41.23$\pm$15.55 & 53.95$\pm$30.24 & 41.23$\pm$15.55 \\
 No Soft Acc & 70.19$\pm$6.86 & 66.25$\pm$13.08 & 71.06$\pm$21.80 & 39.21$\pm$16.17 & 71.06$\pm$21.80 & 39.21$\pm$16.17 \\
 MT-ProtBERT & 
 74.34$\pm$2.15 & 
 74.90$\pm$5.89 & 
 74.07$\pm$3.69 &
 48.76$\pm$4.27 &
 74.07$\pm$3.69 &
 48.76$\pm$4.27 \\
\hline
\end{tabular}}
\label{tab: SApp_1_Abl_Tsk2_Exp3}
\end{table}

%% file: Tables/SApp_Ablt_Tsk2_Exp4.tex
\begin{table}[ht]
\centering
\caption{
Detailed ablation study results comparing model variants for Experiment 4. 
MT-ProtBERT consistently achieves the highest mean Accuracy and MCC across all subsets, validating the necessity of each proposed module. 
Results confirm that Dynamic Window Masking, MS-Conv1D, and multi-task learning jointly drive the performance gain.
}
\resizebox{8.5cm}{!}{
\begin{tabular}{lcccccc}
\hline
 \multirow{2}{1cm}{Setting} & \multicolumn{6}{c}{New dataset 530} \\
\cline{2-7}
 & Accuracy (\%) & F1 Score (\%) & AUPRC (\%) & Sensitivity (\%) & Specificity (\%) & MCC (\%) \\
\hline
 MT-ProtBERT (Mask=1) & 73.58 $\pm$ 5.17 & 72.94 $\pm$ 17.96 & 70.70 $\pm$ 17.81 & 45.26 $\pm$ 11.97 & 70.70 $\pm$ 17.81 & 45.26 $\pm$ 11.97 \\
 ProtBERT Only & 70.19$\pm$2.46 & 56.80$\pm$15.02 & 80.74$\pm$5.34 & 38.63$\pm$10.66 & 80.74$\pm$5.34 & 38.63$\pm$10.66 \\
 No (MLM+STP) & 70.56$\pm$6.35 & 70.17$\pm$9.29 & 70.20$\pm$13.13 & 41.04$\pm$12.83 & 70.20$\pm$13.13 & 41.04$\pm$12.83 \\
 No MLM & 68.68$\pm$6.34 & 67.62$\pm$19.40 & 70.10$\pm$11.26 & 38.93$\pm$12.48 & 70.10$\pm$11.26 & 38.93$\pm$12.48 \\
 No SCD & 71.32$\pm$6.86 & 68.89$\pm$6.08 & 73.59$\pm$14.54 & 42.97$\pm$13.38 & 73.59$\pm$14.54 & 42.97$\pm$13.38 \\
 No MS-Conv1D & 69.43$\pm$4.30 & 70.36$\pm$10.14 & 67.81$\pm$3.13 & 38.04$\pm$9.62 & 67.81$\pm$3.13 & 38.04$\pm$9.62 \\
 No (MS-Conv1D+MLM) & 70.57$\pm$5.91 & 65.93$\pm$14.57 & 75.47$\pm$18.71 & 43.00$\pm$11.41 & 75.47$\pm$18.71 & 43.00$\pm$11.41 \\
 No (MS-Conv1D+STP) & 73.21$\pm$7.71 & 78.55$\pm$9.15 & 67.59$\pm$15.00 & 46.37$\pm$16.13 & 67.59$\pm$15.00 & 46.37$\pm$16.13 \\
 No $AA_{2}$ & 73.96$\pm$4.70 & 74.84$\pm$18.89 & 70.20$\pm$12.12 & 47.22$\pm$9.84 & 70.20$\pm$12.12 & 47.22$\pm$9.84 \\
 No Soft Acc & 73.58$\pm$4.01 & 71.87$\pm$14.30 & 72.24$\pm$18.53 & 46.49$\pm$8.60 & 72.24$\pm$18.53 & 46.49$\pm$8.60 \\
 MT-ProtBERT & 77.74$\pm$3.10 & 79.53$\pm$13.44 & 75.55$\pm$14.22 & 56.79$\pm$5.57 & 75.55$\pm$14.22 & 56.79$\pm$5.57 \\
\hline
\end{tabular}}
\label{tab: SApp_1_Abl_Tsk2_Exp4}
\end{table}

%% file: 2_Appendix_Prove_2AA_is_better.tex
MT-ProtBERT predicts both AA1 and AA2 in the MLM task, which constitutes a multi-view approach. To prove the advantage of training MLM in a multi-view setting, we provide the following proof.

\input{Tables/SApp_2_multi-view_symbol}

First, Table~\ref{tab: SApp_2_multi_AAs_symbol} shows the definition of all relevant symbols.

The optimization objective of single-view MLM is Eq.~\ref{eq: SApp_1_E_single-view}, while that of multi-view MLM is Eq.~\ref{eq: SApp_2_E_multi-view}. The goal is to prove $\mathcal{L}_{\text{multi}}$ $\leq$ $\mathcal{L}_{\text{single}}$.

\begin{align}
\begin{split}
\mathcal{L}_{\text{single}} &= \mathbb{E}_{M\sim \mathcal{M}}\mathbb{E}_{X\sim \text{DWM}(S,M)} \\
 &\:[-\log p_{\theta}(AA_{m}|h_{\theta}(X))] \label{eq: SApp_1_E_single-view}
\end{split} \\
\begin{split}
\mathcal{L}_{\text{multi}} &= \mathbb{E}_{M\sim \mathcal{M}}\mathbb{E}_{X_{1},\dotsm, X_{K}\overset{iid}{\sim} \text{DWM}(S,M)} \\
&\:[-\log(\frac{1}{K}\displaystyle \sum^{K}_{i=1} p_{\theta}(AA_{m}|h_{\theta}(X_{i})))] \label{eq: SApp_2_E_multi-view}
\end{split}
\end{align}

We apply Jensen's inequality to Eq.~\ref{eq: SApp_2_E_multi-view} to obtain

\begin{align}
\begin{split}
Eq.~\ref{eq: SApp_2_E_multi-view} &\leq \mathbb{E}_{M\sim \mathcal{M}}\mathbb{E}_{X_{1},\dotsm, X_{K}\overset{iid}{\sim} \text{DWM}(S,M)} \\
 &\quad [\frac{1}{K}\displaystyle \sum^{K}_{i=1}-\log(p_{\theta}(AA_{m}|h_{\theta}(X_{i})))]
\end{split}
\end{align}

\begin{align}
\begin{split}
 &= \mathbb{E}_{M\sim \mathcal{M}}\frac{1}{K}\displaystyle \sum^{K}_{i=1}\mathbb{E}_{X_{1},\dotsm, X_{K}\overset{iid}{\sim} \text{DWM}(S,M)} \\
 &\: [-\log(p_{\theta}(AA_{m}|h_{\theta}(X_{i})))] \label{eq: SApp_3_Jensen_Ineq}
\end{split}
\end{align}

Next, it is important to prove that

\begin{align}
\begin{split}
\mathbb{E}_{X_{1},\dotsm, X_{K}\overset{iid}{\sim} \text{DWM}(S,M)}[-\log(p_{\theta}(AA_{m}|h_{\theta}(X_{i})))] \\
= \mathbb{E}_{X_{i}\sim \text{DWM}(S,M)}[-\log(p_{\theta}(AA_{m}|h_{\theta}(X_{i})))] \\ \label{eq: SApp_4_reduce_dimension}
\end{split}
\end{align}

Given that $(X_{1},\dotsm, X_{K})$ $\overset{iid}{\sim}$ $\text{DWM}(S,M)$, we have $p(X_{1},\dotsm, X_{K})$ $=$ $\displaystyle \Pi^{K}_{j=1}p(X_{j})$.

Therefore, 

\begin{align}
\begin{split}
Eq.~\ref{eq: SApp_4_reduce_dimension} &= \displaystyle \sum_{X_{1}\in \chi}\dotsm \displaystyle \sum_{X_{K}\in \chi} [-\log(p_{\theta}(AA_{m}|h_{\theta}(X_{i})))] \\
&\quad p(X_{1},\dotsm, X_{K}) \\
\end{split}
\end{align}

\begin{align}
\begin{split}
&= \displaystyle \sum_{X_{1}\in \chi}\dotsm \displaystyle \sum_{X_{K}\in \chi}[-\log(p_{\theta}(AA_{m}|h_{\theta}(X_{i})))] \\
&\quad \displaystyle \Pi^{K}_{j=1}p(X_{j}) \\
&= \displaystyle \sum_{X_{i}\in \chi}[-\log(p_{\theta}(AA_{m}|h_{\theta}(X_{i})))]p(X_{i}) \\
&\quad \displaystyle \Pi_{j\neq i}\displaystyle \sum_{X_{j}}p(X_{j}) \\
&= \displaystyle \sum_{X_{i}\in \chi}[-\log(p_{\theta}(AA_{m}|h_{\theta}(X_{i})))] \\
&\quad p(X_{i})\displaystyle \Pi_{j\neq i} 1 \\
&= \displaystyle \sum_{X_{i}\in \chi}[-\log(p_{\theta}(AA_{m}|h_{\theta}(X_{i})))]p(X_{i}) \\
&= \mathbb{E}_{X_{i}\sim \text{DWM}(S,M)}[-\log(p_{\theta}(AA_{m}|h_{\theta}(X_{i})))]
\end{split} \label{eq: SApp_5_multi=single}
\end{align}

Now, Eq.~\ref{eq: SApp_5_multi=single} proves Eq.~\ref{eq: SApp_4_reduce_dimension}. We insert Eq.~\ref{eq: SApp_5_multi=single} into Eq.~\ref{eq: SApp_3_Jensen_Ineq}.

\begin{align}
\begin{split}
Eq.~\ref{eq: SApp_3_Jensen_Ineq} &= \mathbb{E}_{M\sim \mathcal{M}}\frac{1}{K}\displaystyle \sum^{K}_{i=1}\mathbb{E}_{X_{i}\sim \text{DWM}(S,M)} \\
 &\quad [-\log(p_{\theta}(AA_{m}|h_{\theta}(X_{i})))] \\
 &= \mathbb{E}_{M\sim \mathcal{M}}\frac{1}{K}\displaystyle \sum^{K}_{i=1} \mathcal{L}_{\text{single}} \\
 &= \mathbb{E}_{M\sim \mathcal{M}} \mathcal{L}_{\text{single}} \\
 &= \mathcal{L}_{\text{single}}
\end{split}
\end{align}

Hence, we prove $\mathcal{L}_{\text{multi}} \leq \mathcal{L}_{\text{single}}$.

\textbf{When $\mathcal{L}_{\text{multi}}$ $=$ $\mathcal{L}_{\text{single}}$?}

The prediction probabilities of all views are identical to each other.

$p_{\theta}(AA_{m}|h_{\theta}(X_{1}))$ $=$ $\dotsm$ $=$ $p_{\theta}(AA_{m}|h_{\theta}(X_{K}))$

\textbf{When $\mathcal{L}_{\text{multi}}$ $<$ $\mathcal{L}_{\text{single}}$?}

The prediction probabilities of at least two views are unequal.

$p_{\theta}(AA_{m}|h_{\theta}(X_{1}))$ $\neq$ $\dotsm$ $\neq$ $p_{\theta}(AA_{m}|h_{\theta}(X_{K}))$

%% file: Tables/SApp_2_multi-view_symbol.tex
\begin{table}[ht]
\centering
\caption{Symbols appeared in the proving the advantage of doing multi-view MLM.}
\resizebox{8cm}{!}{
\begin{tabular}{c|l}
\hline
 Symbol & Description\\
\hline
 S & Grounth truth sequence \\
 DWM & Dynamic Windows Masking Method \\
 iid & Independent and identical distribution \\
 \multirow{3}{*}{$X_{1},\dotsm, X_{K}$} & \multirow{3}{7cm}{Multi-view after different masking. $X_{i} \bot X_{j}|S$, $i,j\in$ $[1,K]$ and $i\neq j$. They are conditionally independent.} \\
 & \\ & \\
 $h_{\theta}(\cdot)$ & Shared ProtBERT backbone \\
 $AA_{m}$ & Masked amino acids by method m \\
 \multirow{2}{*}{$p_{\theta}(AA_{m}|h_{\theta}(X_{i}))$} & \multirow{2}{7cm}{Model's predicted probability for the masked AA given view $X_{i}$.} \\
 & \\
 \multirow{4}{*}{$\mathbb{E}_{M\sim \mathcal{M}}$} & \multirow{4}{7cm}{Expectation of all masking methods. It optimizes, under all possible missing modes, the average recovering capability of masked amino acids.} \\
 & \\ & \\ & \\
 \multirow{2}{*}{$\mathbb{E}_{X_{i}\sim \text{DWM}(S,M)}$} & \multirow{2}{7cm}{Given certain sequence and masking method, the probability of correct prediction.} \\
 & \\
\hline
\end{tabular}}
\label{tab: SApp_2_multi_AAs_symbol}
\vspace{-0.5cm}
\end{table}